\documentclass[10pt,twocolumn,letterpaper]{article}

\usepackage{multirow}
\usepackage{makecell}
\usepackage[pagenumbers]{iccv} 

\DeclareUnicodeCharacter{FF0C}{,}

\definecolor{iccvblue}{rgb}{0.21,0.49,0.74}
\usepackage{subcaption}
\usepackage[pagebackref,breaklinks,colorlinks,allcolors=iccvblue]{hyperref}

\def\paperID{*****} 
\def\confName{ICCV}
\def\confYear{2025}

\title{PP-DocLayout: A Unified Document Layout Detection Model to Accelerate Large-Scale Data Construction}

\author{Ting Sun, Cheng Cui, Yuning Du, Yi Liu \\
\textsuperscript{} PaddlePaddle Team, Baidu Inc.\\
\tt\small \{sunting13, cuicheng01\} @baidu.com
}

\begin{document}
\maketitle
\begin{abstract}
Document layout analysis is a critical preprocessing step in document intelligence, enabling the detection and localization of structural elements such as titles, text blocks, tables, and formulas. Despite its importance, existing layout detection models face significant challenges in generalizing across diverse document types, handling complex layouts, and achieving real-time performance for large-scale data processing. To address these limitations, we present PP-DocLayout, which achieves high precision and efficiency in recognizing 23 types of layout regions across diverse document formats. To meet different needs, we offer three models of varying scales. \textbf{PP-DocLayout-L} is a high-precision model based on the RT-DETR-L detector, achieving 90.4\% mAP@0.5 and an end-to-end inference time of 13.4 ms per page on a T4 GPU. \textbf{PP-DocLayout-M} is a balanced model, offering 75.2\% mAP@0.5 with an inference time of 12.7 ms per page on a T4 GPU. \textbf{PP-DocLayout-S} is a high-efficiency model designed for resource-constrained environments and real-time applications, with an inference time of 8.1 ms per page on a T4 GPU and 14.5 ms on a CPU. This work not only advances the state of the art in document layout analysis but also provides a robust solution for constructing high-quality training data, enabling advancements in document intelligence and multimodal AI systems. Code and models are
available at \url{https://github.com/PaddlePaddle/PaddleX}.
\end{abstract}    
\section{Introduction}
\label{sec:intro}

\begin{figure*}[t]
    \centering
    \includegraphics[width=\linewidth]{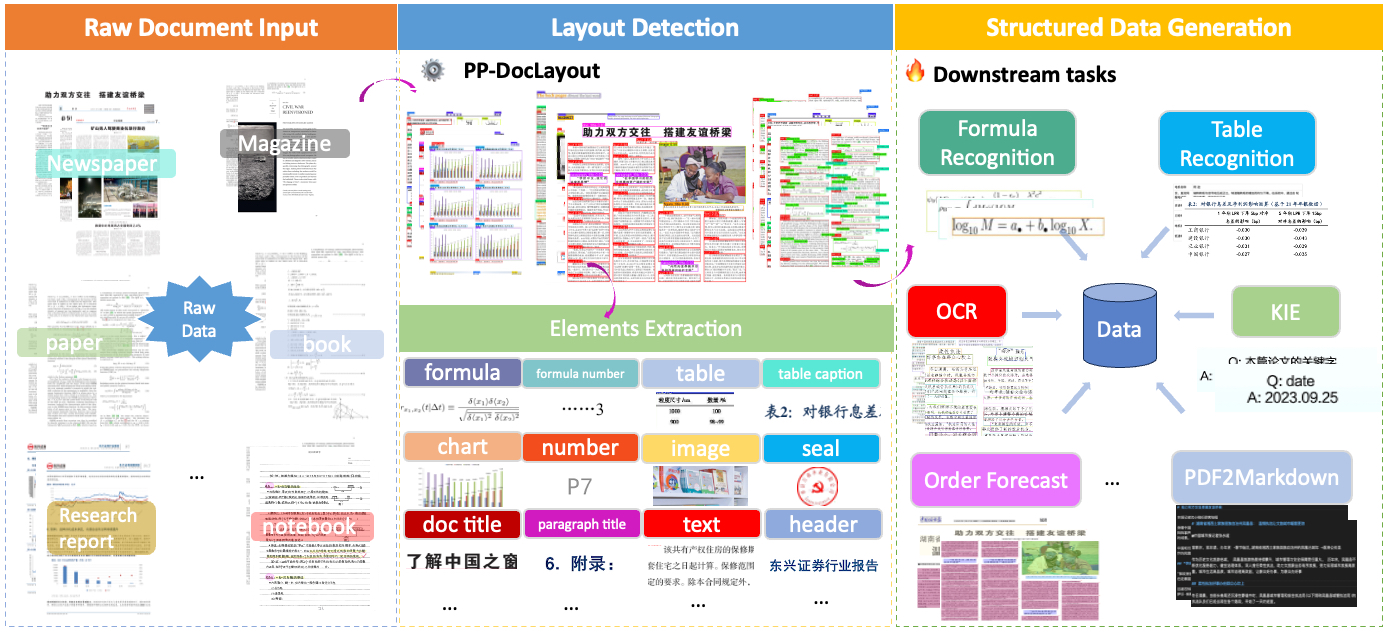}
    \caption{PP-DocLayout plays a pivotal role in layout analysis and structured data generation.}
    \label{fig:pipeline}
\end{figure*}

The rapid development of large language models (LLMs) and multi-modal document understanding systems \cite{xu2020layoutlm,xu2020layoutlmv2} has led to a significant increase in the demand for high-quality structured training data.  Document layout detection, which identifies and localizes structural elements (e.g., text blocks, tables, and figures), plays a pivotal role in transforming raw document images into machine-readable formats. As illustrated in \textbf{Figure~\ref{fig:pipeline}}, layout detection is the foundational step for a variety of downstream tasks, including table recognition, formula recognition, OCR and information extract. For instance, in the case of table recognition, layout detection models accurately locate and define the boundaries of tables within document images, enabling the extraction of table regions for further processing, such as parsing the tabular structure and extracting the underlying data. This structured table data is extremely valuable for applications ranging from data analysis to information retrieval. Similarly, for formula recognition, layout detection models detect and localize formula regions within documents. This allows for the extraction of these regions, which can then be fed into specialized formula recognition systems. The resulting structured formula data not only enhances the machine's understanding of mathematical content but also enriches training datasets, improving models' ability to recognize and interpret formulas in various contexts. 

However, despite its potential, existing layout detection models face three critical limitations. (1) Poor generalization across document types. Current approaches predominantly focus on academic papers, resulting in poor performance on other document types such as magazines, newspapers, and financial reports. (2) Inadequate handling of complex layouts. The lack of comprehensive category definitions—such as the absence of separate labels for inline and interline mathematical formulas—necessitates auxiliary models, leading to increased complexity and reduced efficiency. (3) Insufficient processing speed for real-time applications. These challenges impede the effective use of layout detection in practical scenarios, particularly in domains where training large models requires efficient acquisition of abundant massive high-quality data. 

To address these challenges, we introduce PP-DocLayout, a unified layout detection model that achieves state-of-the-art accuracy and real-time inference capabilities. PP-DocLayout achieves high-precision recognition and localization of layout regions across diverse document formats including Chinese or English academic papers, research reports, exam papers, books, newspapers, and magazines. This advancement significantly enhances the diversity and quality of large models training data acquisition.  In addition, PP-DocLayout series supports 23 common layout categories, which cover a wide range of layout elements found in diverse documents. This clear hierarchical structure of categories facilitates improved semantic understanding and logical parsing, while the inclusion of structured data of high-value information enables more precise data processing and analysis. For addressing the critical efficiency requirements for massive-scale data construction, leveraging the high-performance PaddleX inference engine, the lightweight model demonstrates exceptional processing capabilities - handling approximately 123 pages per second on T4 GPU. These performance substantially outperforms existing open-source solutions, establishing new benchmarks for document layout analysis in terms of both accuracy and computational efficiency.

\section{Related Work}
\label{sec:formatting}

The evolution of document layout analysis (DLA) reflects the paradigm shift from isolated component detection to holistic semantic understanding. Early unimodal methodologies framed DLA as a specialized computer vision task, adapting generic object detection frameworks (Faster R-CNN \cite{ren2016faster}, YOLO \cite{wang2024yolov10}) with domain-specific modifications. Recently, the advanced method DocLayout-YOLO \cite{zhao2024doclayout} ，based on YOLOv10 \cite{wang2024yolov10}， pretrains on diverse document data and designs the GL-CRM module which obtains high-accuracy of 10 categories layout detection. 

The emergence of multimodal learning has fundamentally transformed DLA methodology.  The LayoutLM series \cite{xu2020layoutlm,xu2020layoutlmv2,huang2022layoutlmv3} demonstrated the power of unified pre-training strategies, integrating masked visual-language modeling and spatial-aware position embeddings. Recent advancements further explore self-supervised paradigms, DiT \cite{li2022dit} leverages massive unlabeled documents through novel pre-training objectives, and VGT \cite{da2023vision} introduces grid-based textual encoding to preserve fine-grained typographical features. Notably, the field is witnessing convergence between DLA and document intelligence, where layout understanding serves as the foundation for higher-level semantic tasks.

Despite these advancements, several challenges remain. First, most existing methods focus on specific document types, such as academic papers, and lack generalization to diverse document categories like magazines, newspapers, and handwritten notes. Second, the detection of fine-grained elements, such as formulas, footnotes, and headers, remains under explored. Finally, the computational efficiency of layout detection methods remains a significant challenge, as many state-of-the-art models are computationally expensive and slow, limiting their applicability in real-time or large-scale document processing scenarios.

Our work addresses these limitations by proposing a unified framework for document layout detection that supports a wide range of document types and fine-grained element categories. By leveraging advanced deep learning techniques and incorporating contextual information, our method achieves robust performance across diverse layouts while maintaining computational efficiency.


\section{Method}

We introduce PP-DocLayout, a unified detection model achieving state-of-the-art performance through innovations in both data curation and algorithm design. Our approach combines three key improvement strategies.

\begin{figure}[t]
    \centering
    \includegraphics[width=\linewidth]{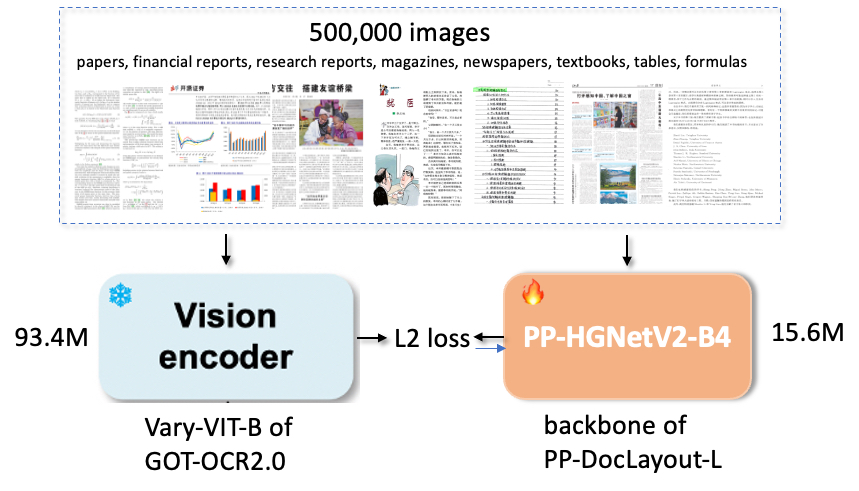}
    \caption{Knowledge Distillation Framework for the backbone of PP-DocLayout-L}
    \label{fig:kl}
\end{figure}

\subsection{Knowledge Distillation Framework} PP-DocLayout-L employs a knowledge distillation \cite{hinton2015distilling} paradigm to enhance the performance of document layout understanding as shown in \textbf{Figure~\ref{fig:kl}}. In this framework, the visual encoder Vary-VIT-B model of GOT-OCR2.0 \cite{wei2024general} acts as the teacher model, which is a well-trained and robust model possessing advanced document understanding capabilities. The student model, in this case, is the PP-HGNetV2-B4 backbone of PP-DocLayout-L, which is designed to learn from the teacher model.

The distillation process involves transferring knowledge from the teacher to the student by aligning their feature representations.
Specifically, The teacher network parameters remain frozen while training PP-HGNetV2-B4, leveraging feature-level supervision through a fully-connected layer. Let $\mathbf{F}_{\text{Vary-VIT-B}} \in \mathbb{R}^{B \times D}$ and $\mathbf{F}_{\text{PP-HGNetV2-B4}} \in \mathbb{R}^{B \times P}$ denote the feature tensors from teacher and student networks respectively, where $D$ and $P$ represent their corresponding feature dimensions, $B$ denotes the Batch size. To bridge the dimensional discrepancy ($D \neq P$), we introduce a learnable linear projection $\varphi: \mathbb{R}^{P} \to \mathbb{R}^{D}$. The distillation loss is formulated as:

\begin{equation} 
\mathcal{L}_{\text{Distill}} = \frac{1}{B} \sum_{i=1}^{B} \left\lVert \mathbf{F}_{\text{Vary-VIT-B}}^{(i)} - \varphi(\mathbf{F}_{\text{PP-HGNetV2-B4}}^{(i)}) \right\rVert_{2}^{2}
\end{equation}

The distillation framework was trained on a diverse corpus containing 500000 document samples spanning five domains: 
\begin{itemize}
    \item Mathematical formulas (including equation derivations and symbolic notations)
    \item Financial documents (reports and balance sheets)
    \item Scientific literature (arxiv papers in STEM fields)
    \item Academic dissertations (with complex layout structures)
    \item Tabular data (statistical reports and spreadsheets)
\end{itemize}

Training was conducted at $768\times$768 resolution for 50 epochs using AdamW optimizer ($\beta_1=0.9$, $\beta_2=0.999$). The distilled PP-HGNetV2-B4 achieves effective feature extraction capabilities with only 15.6 M parameters.

\subsection{Semi-supervised Learning} In this section, we present our semi-supervised learning approach employed to enhance the performance of PP-DocLayout-M and PP-DocLayout-S models. This method leverages the high-precision capabilities of the PP-DocLayout-L model to generate pseudo-labels, which are subsequently used to augment the training data for the less complex models.

\textbf{Pseudo-Label Generation}
Given an unlabeled document image $x_u$, we first generate raw predictions using the teacher model PP-DocLayout-L with trained parameters $\theta_T$ as follow:

\begin{equation}
P(y|x_u) = f_{\text{T}}(x_u; \theta_T)
\end{equation}

where $P(y|x_u) \in \mathbb{R}^{N \times C}$ contains prediction scores for $N$ potential regions across $C=23$ layout categories.

\textbf{Adaptive Threshold Selection}
Traditional fixed-threshold approaches often suffer from class imbalance and difficulty levels of learning in different categories. Thus, we propose an adaptive threshold selection method to obtain high-quality pseudo-labels. Our threshold selection strategy explicitly maximizes the F-score on labeled data $x_l$ through a systematic optimization process. For each layout category $c \in \{1,...,23\}$, we determine the optimal threshold $\tau_c^*$ by solving:

\begin{equation}
\tau_c^* = \mathop{\arg\max}_{\tau \in [0,1]} F_c(\tau; P(y|x_l))
\label{eq:fscore-opt}
\end{equation}

where $F_c(\tau)$ denotes the F1-score computed on the validation set $\mathcal{V}$ when using threshold $\tau$ for class $c$. 

After determining the optimal thresholds for each class, we generate pseudo-labels for the unlabeled data. For each potential region in the document image $x_u$, a pseudo-label is assigned if the prediction score exceeds the corresponding optimal threshold $\tau_c^*$.

\textbf{Pseudo-Label Generation and Training}
Using the optimized thresholds $\{\tau_1^*, \tau_2^*, \ldots, \tau_{23}^*\}$, the pseudo-labels $\hat{y}_u$ for the unlabeled document image $x_u$ are defined as:

\begin{equation}
\hat{y}_{u,i} = 
\begin{cases} 
1, & \text{if } P(y_{i,c}|x_u) > \tau_c^* \text{ for any } c \\
0, & \text{otherwise}
\end{cases}
\end{equation}

where $\hat{y}_{u,i}$ indicates whether the $i$-th region is assigned a pseudo-label for any category.
The pseudo-labeled data, along with the real labeled data, form a comprehensive training set that enhances the learning process of the student models PP-DocLayout-S and PP-DocLayout-M. By integrating the high-quality pseudo-labels with the labeled data, the models can generalize better, learning robust features that improve document layout detection.


\section{Experimental Results}

\begin{table}[t]
    \centering
    \small 
    \begin{tabular}{|c|c|}
        \hline
        \textbf{Our Method} &  \textbf{DocLayout-YOLO} \\
        \hline
        Document Title & Title \\
        Paragraph Title  & Title \\
        Text  & Plain Text \\
        Page Number & Abandon \\
        Abstract  & Plain Text \\
        Content & Plain Text \\
        Reference  & Plain Text \\
        Footnote &  Abandon \\
        Header  & Abandon \\
        Footer & Abandon \\
        Algorithm  & Plain Text \\
        Formula  &  Isolate Formula \\
        Formula Number  & Formula Number \\
        Image  & Figure \\
        Image Caption  & Figure Caption \\
        Table  & Table \\
        Table Caption  & Table Caption \\
        Chart  & Figure \\
        Chart Caption & Figure Caption \\
        Seal  & Figure \\
        Header Image  & Abandon \\
        Footer Image  & Abandon \\
        Aside Text  & Abandon \\
        \hline
    \end{tabular}
    \caption{Comparison of categories recognized by our method and another advanced algorithm DocLayout-YOLO \cite{zhao2024doclayout}.}
    \label{tab:category_comparison}
\end{table}

\begin{table*}[h]
    \centering
    \begin{tabular}{lcccc}
        \toprule
        \textbf{Model Name} & \textbf{mAP@0.5 (\%)} & \textbf{GPU Inference (ms)} & \textbf{CPU Inference (ms)} & \textbf{Params (M)} \\
        \midrule
        PP-DocLayout-L & 90.4 & 13.39 & 759.76 & 30.94 \\
        PP-DocLayout-M & 75.2 & 12.73 & 59.82 & 5.65 \\
        PP-DocLayout-S & 70.9 & 8.11 & 14.49 & 1.21 \\
        \bottomrule
    \end{tabular}
    \caption{Model performance and specifications. \noindent\textit{Note:} GPU inference times are based on an NVIDIA Tesla T4 machine. CPU inference speeds are based on an Intel(R) Xeon(R) Gold 6271C CPU @ 2.60GHz, with 8 threads and FP16 precision.}
    \label{tab:model_performance}
\end{table*}
 
\subsection{Dataset} we collected a comprehensive dataset of document images encompassing a wide variety of types such as Chinese and English academic papers, magazines, newspapers, research reports, exam papers, handwritten notes, contracts, and books. This diverse dataset ensures the robustness and generalizability of our model across different document formats and structures. The dataset comprises 30,000 images for training and 500 images for evaluation. Images are collected from Baidu image search and public datasets, including Doclaynet \cite{pfitzmann2022doclaynet} and PublayNet \cite{zhong2019publaynet}. Images are annotated with 23 common layout categories and the distribution of these categories is detailed in the Table \ref{tab:instance_counts} in the appendix. 

As shown in \textbf{Table~\ref{tab:category_comparison}}, our method defines a more comprehensive and fine-grained set of categories compared to DocLayout-YOLO \cite{zhao2024doclayout}. While DocLayout-YOLO simplifies many document elements into broad classes such as ``title,'' ``text,'' and ``figure,'' our method distinguishes between semantically meaningful elements like document titles, paragraph titles, page numbers, headers, footers, and footnotes. This granularity enables better parsing of the document's hierarchical structure and logical relationships. Additionally, our method accurately identifies and categorizes high-value elements such as formulas, charts, and seals, which DocLayout-YOLO either misclassifies or ignores (e.g., labeling them as ``abandon'' or ``figure''). This comprehensive categorization supports a wider range of downstream tasks, including document understanding, information extraction, and format conversion.

\subsection{Implementation Details}

The \textbf{PP-DocLayout-L} model is built upon the RT-DETR-L \cite{zhao2024detrs} object detection architecture and utilizes a pre-trained PPHGNetV2-B4 model that has undergone knowledge distillation. The training was configured with a constant learning rate of 0.0001. The model was trained for 100 epochs with a batch size of 2 per GPU, using a total of 8 GPUs, resulting in a total training time of approximately 26 hours on NVIDIA V100 GPUs.
The \textbf{PP-DocLayout-M} and \textbf{PP-DocLayout-S} models are based on the PicoDet-M and PicoDet-S \cite{yu2021pp} object detection architecture, respectively. Both models were trained for 100 epochs with a batch size of 2 per GPU, utilizing a total of 8 GPUs. The learning rates were set to for 0.02 PP-DocLayout-M and 0.06 for PP-DocLayout-S, and were dynamically adjusted using the CosineDecay \cite{loshchilov2016sgdr} learning rate scheduler.

\subsection{Main Results}

\textbf{Table~\ref{tab:model_performance}} presents the performance and specifications of different variants of the PP-DocLayout model. The table highlights the trade-offs between accuracy, inference speed, and model size for each model variant.

The PP-DocLayout-L model achieves the highest accuracy with a mean Average Precision (mAP) of 90.4\% at an IoU threshold of 0.5. However, this accuracy comes with a model size of 30.94 million parameters and inference times, taking 13.39 milliseconds on a T4 GPU, about 74.6 FPS and approximately 759.76 milliseconds on a CPU, which translates to about 1.32 FPS. Referring to \textbf{Figure~\ref{fig:visualizations}} in the appendix, we provide additional visualization results to further demonstrate the effectiveness of our model across a diverse range of document types and layouts. Specifically, we visualize the performance of our method on documents such as \textit{papers}, \textit{magazines}, \textit{newspapers}, \textit{research reports}, \textit{books}, \textit{notebooks}, \textit{contracts} and \textit{test papers}. The visualizations clearly show that our model accurately identifies and categorizes diverse elements.

The PP-DocLayout-S model offers a significantly smaller model size of 1.21 million parameters with faster inference times 8.11 milliseconds on a T4 GPU, about 123 FPS and 14.49 milliseconds on a CPU, which translates to approximately 69.04 FPS. Despite its compactness, it maintains a respectable mAP of 70.9\%.

The PP-DocLayout-M model strikes a balance between the two extremes, achieving an mAP of 75.2\%. It has a moderate model size of 5.65 million parameters, with inference times of 12.73 milliseconds on a T4 GPU and about 59.82 milliseconds on a CPU, translating to approximately 16.72 FPS.

These results illustrate the trade-offs inherent in model design, where increases in accuracy often come at the expense of model size and inference speed. The choice of model may thus depend on the specific requirements of accuracy, computational resources, and latency in practical applications.

\subsection{Qualitative Analysis}

\begin{figure*}[htbp]
\centering
\begin{tabular}{c c c c c}
    \parbox[c]{1.23cm}{\textbf{DocLayout\\YOLO}} & 
    \includegraphics[width=0.2\textwidth]{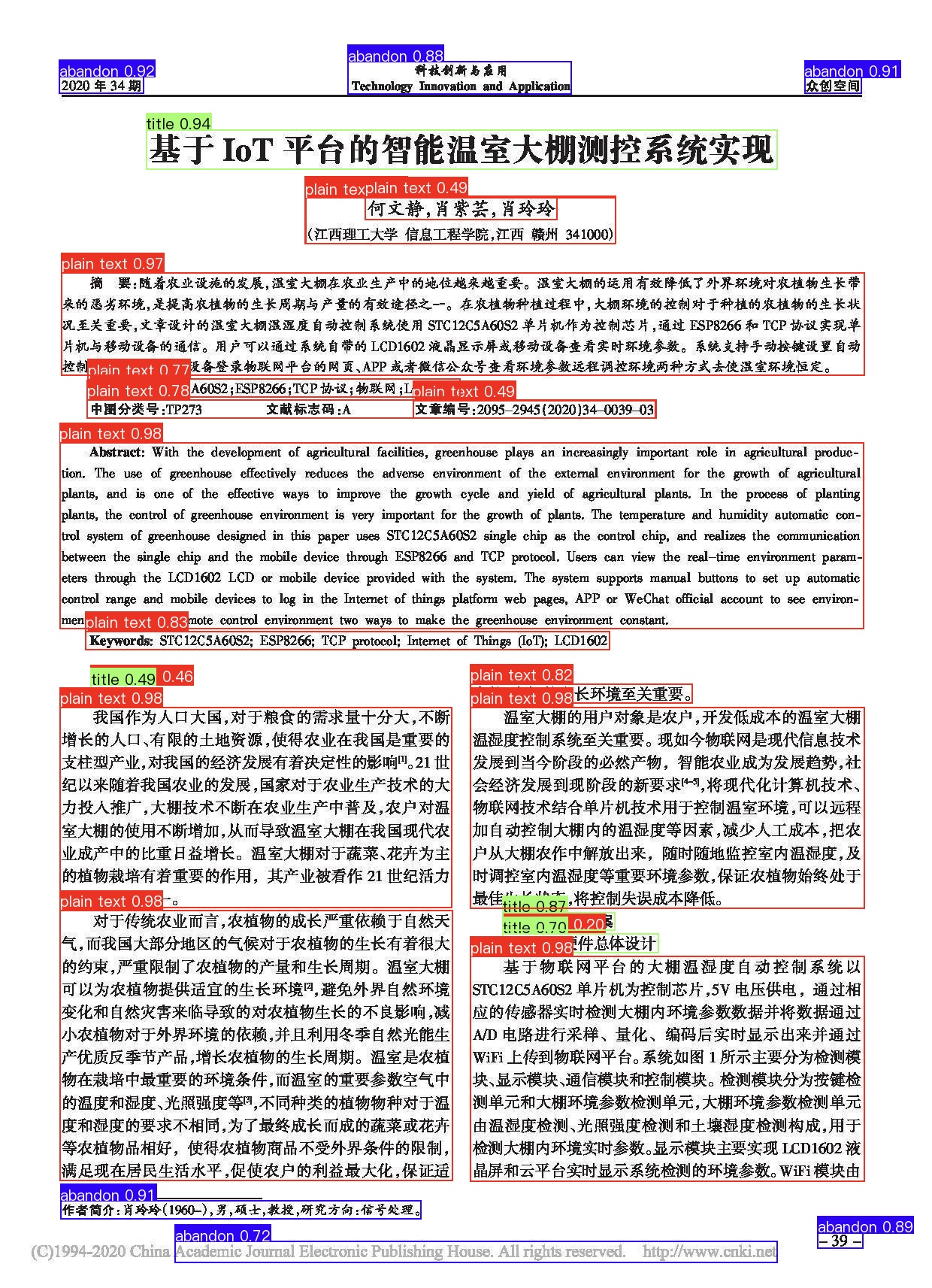} & 
    \includegraphics[width=0.2\textwidth]{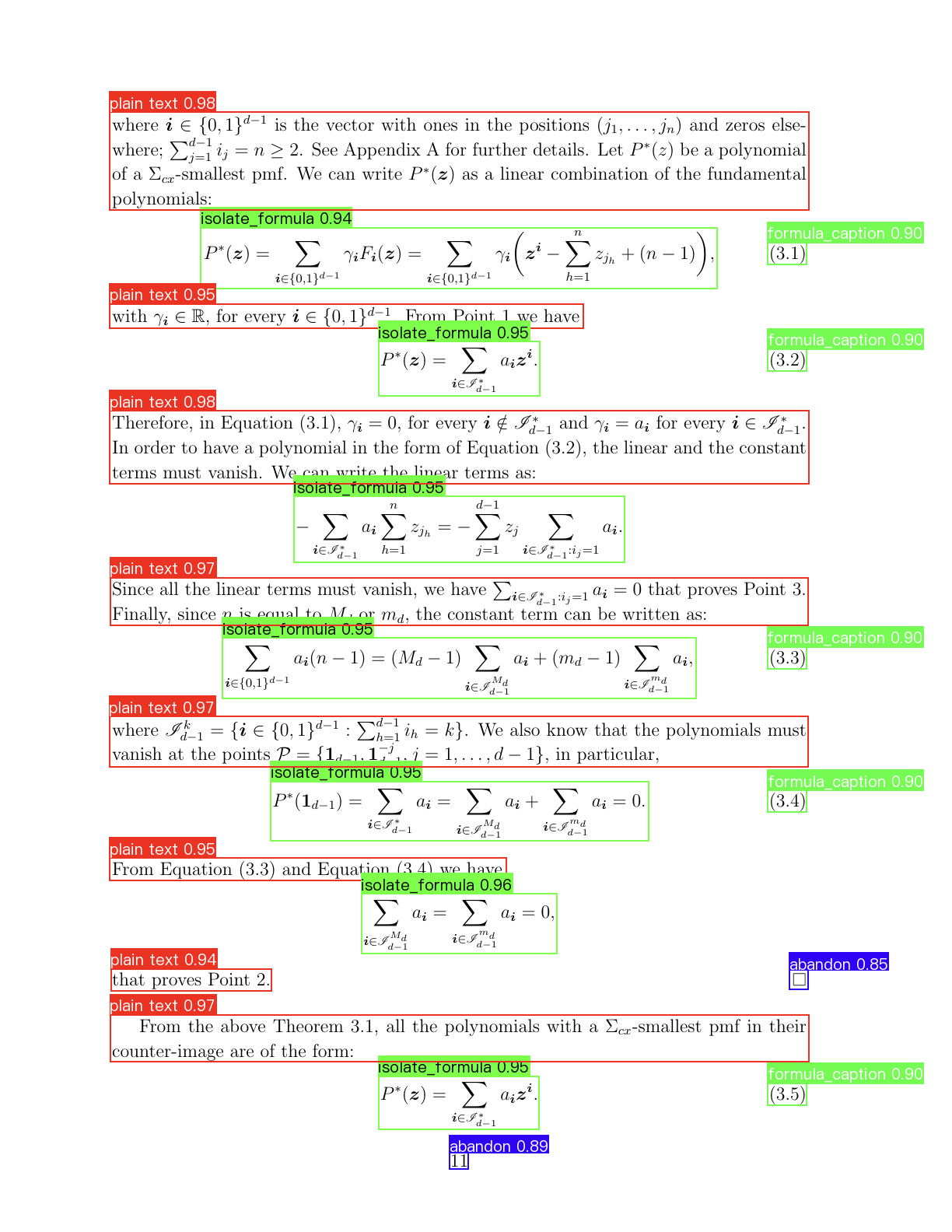} & 
    \includegraphics[width=0.2\textwidth]{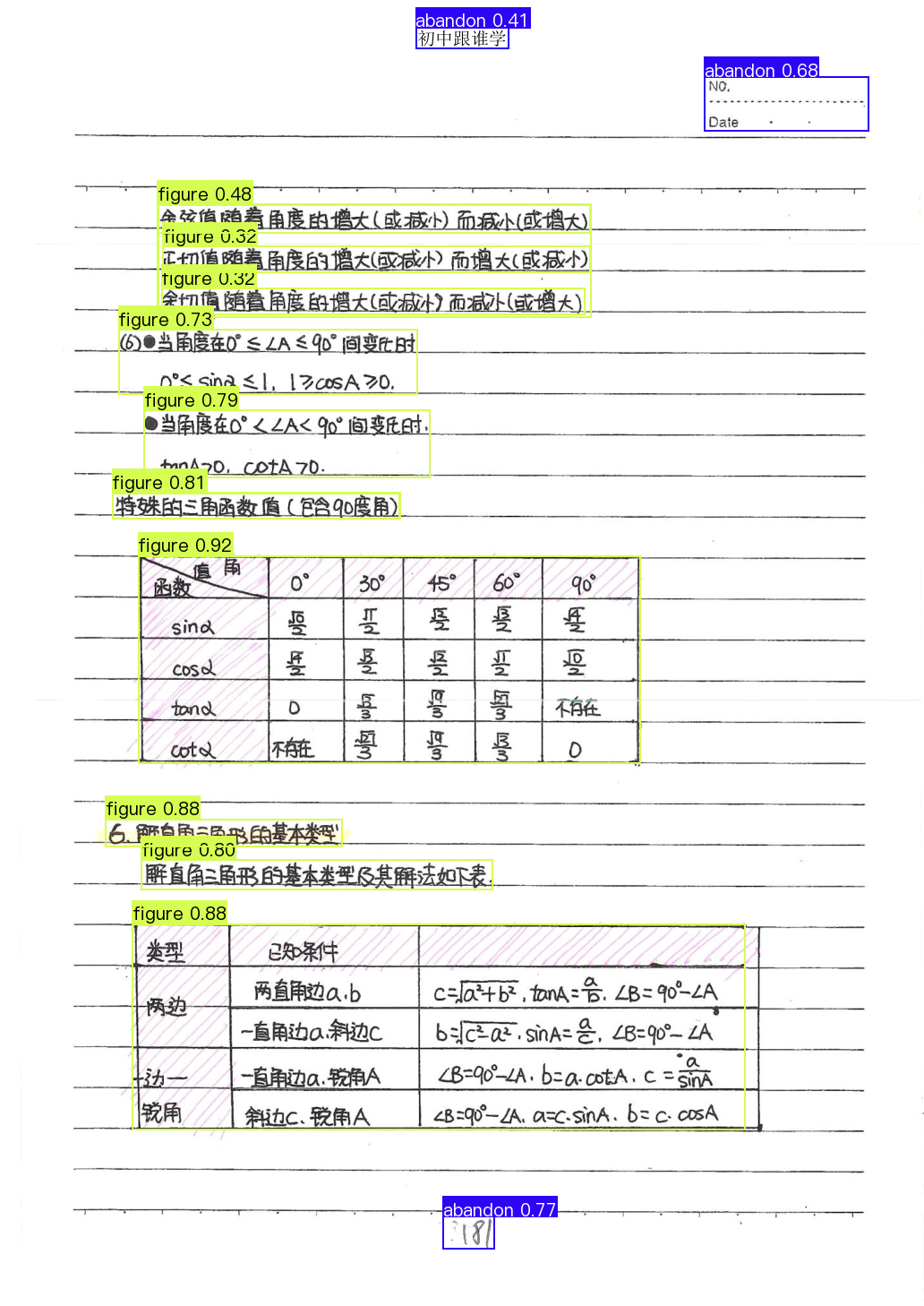} & 
    \includegraphics[width=0.2\textwidth]{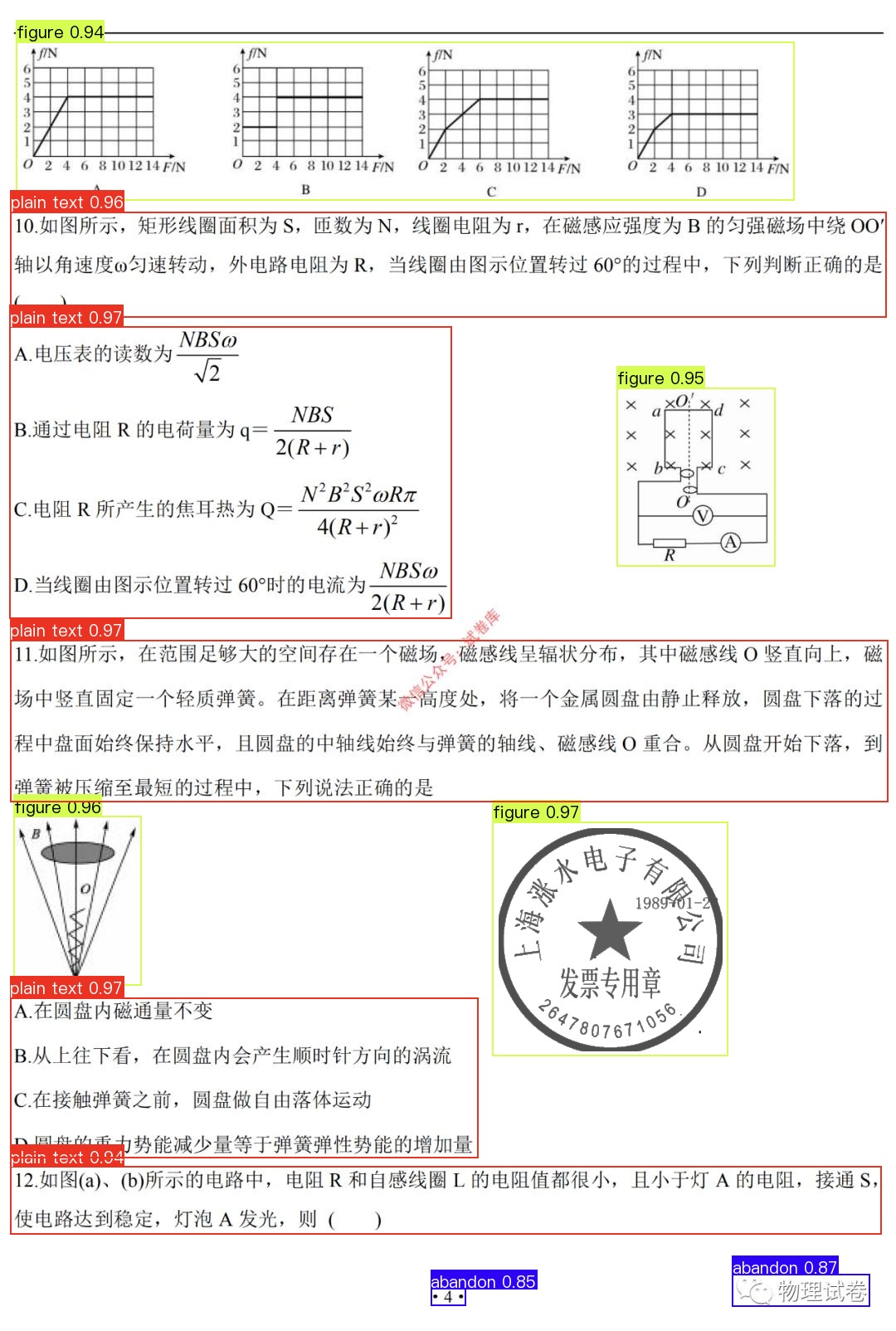} \\
    \parbox[c]{1.23cm}{\textbf{Our\\Results}} & 
    \includegraphics[width=0.2\textwidth]{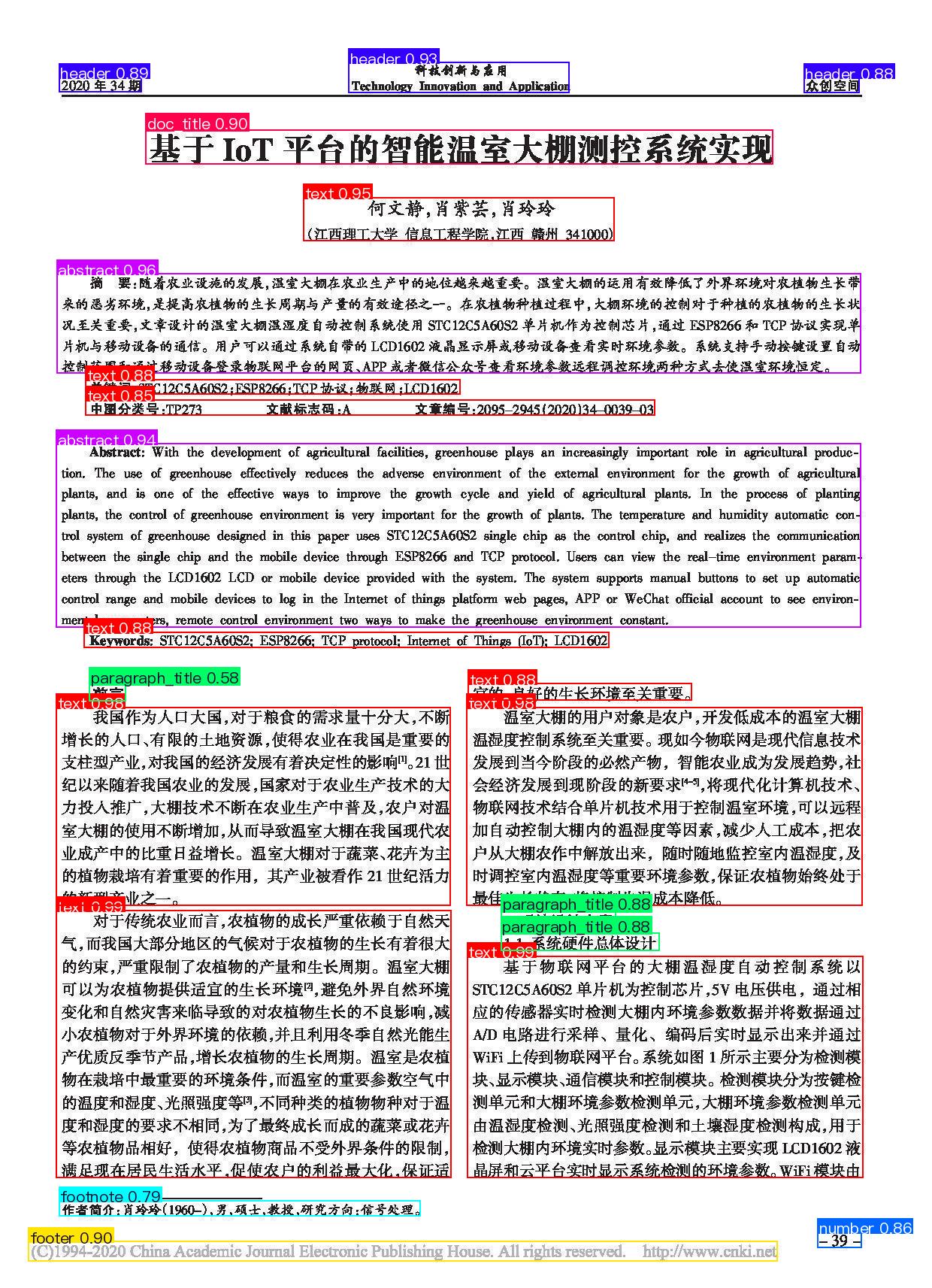} & 
    \includegraphics[width=0.2\textwidth]{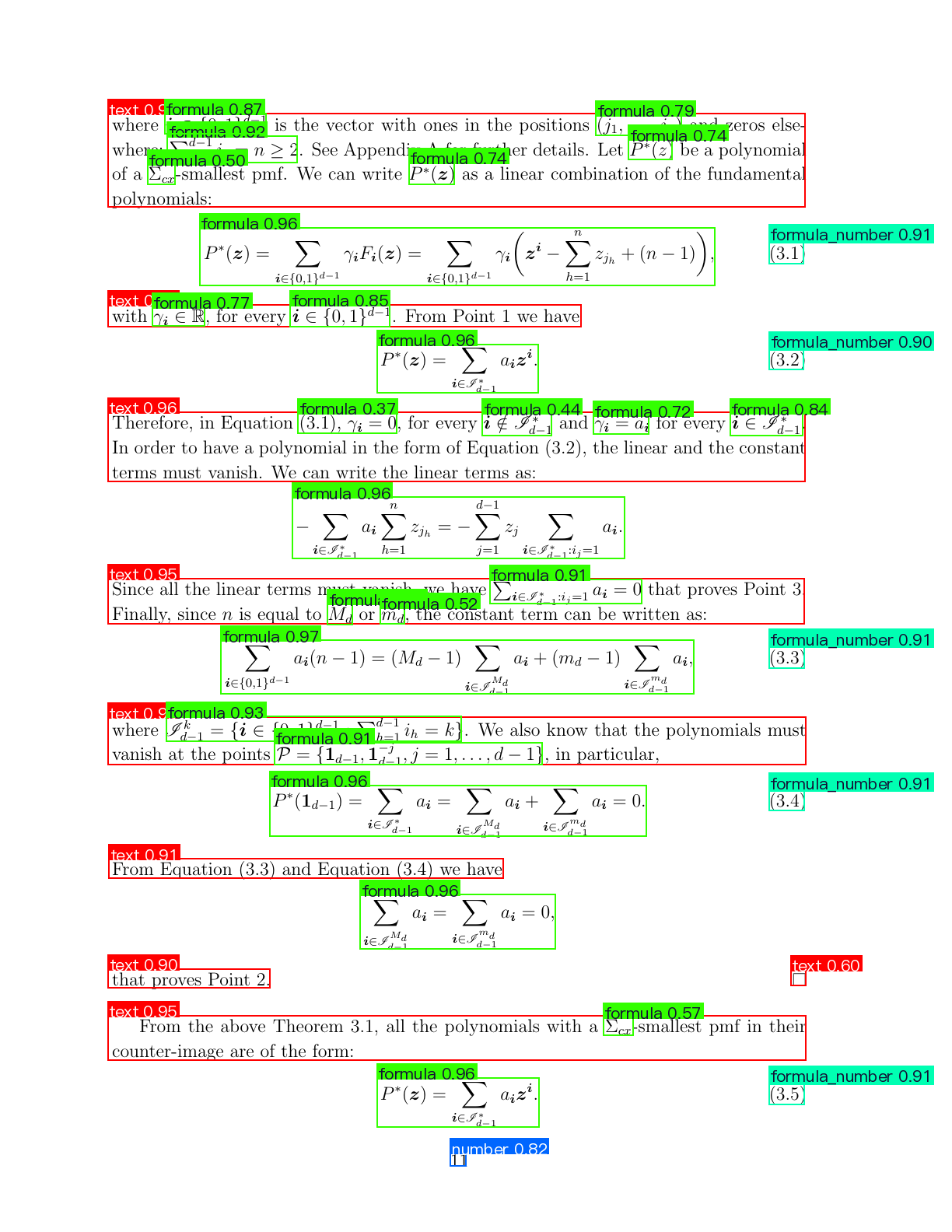} & 
    \includegraphics[width=0.2\textwidth]{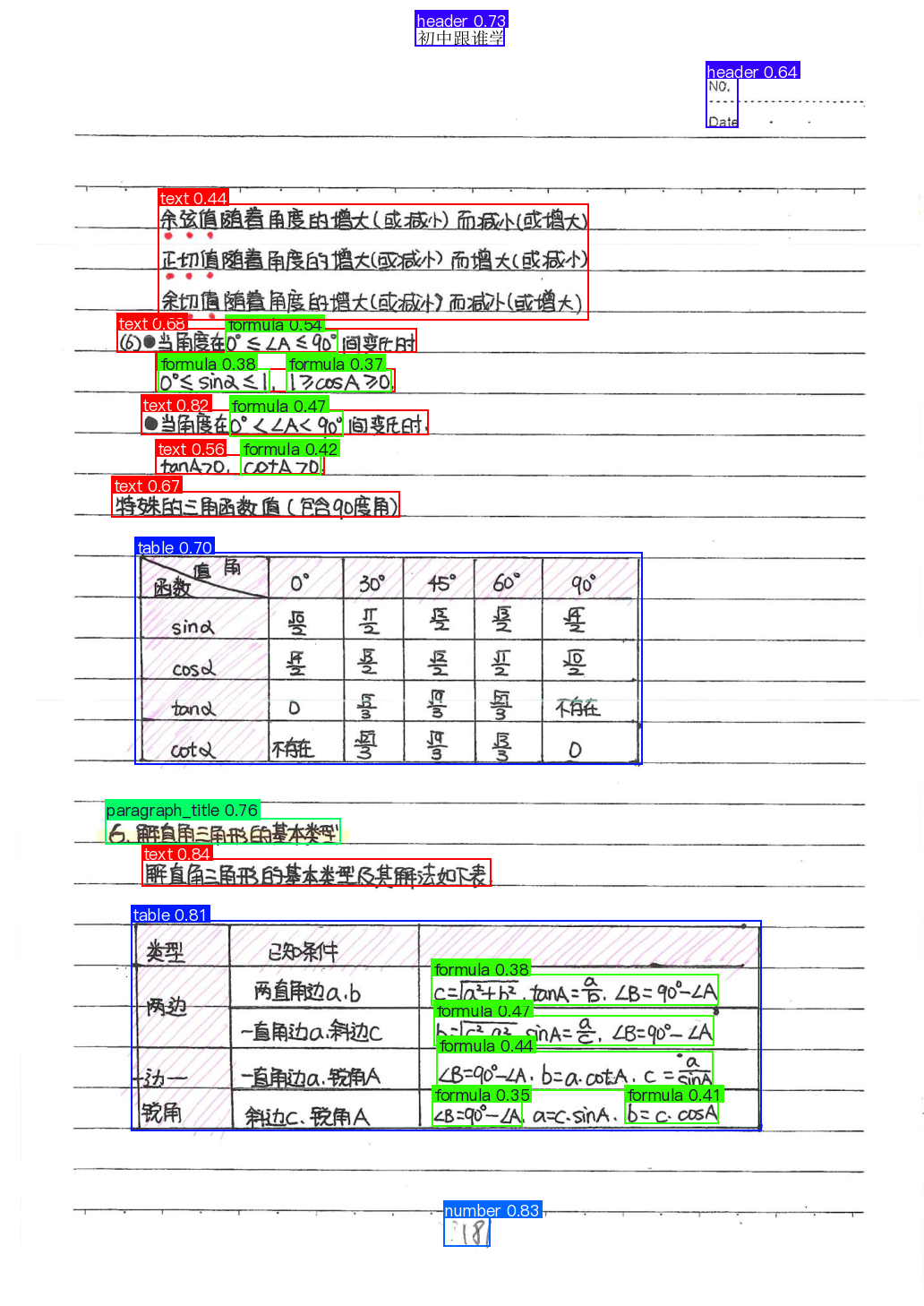} & 
    \includegraphics[width=0.2\textwidth]{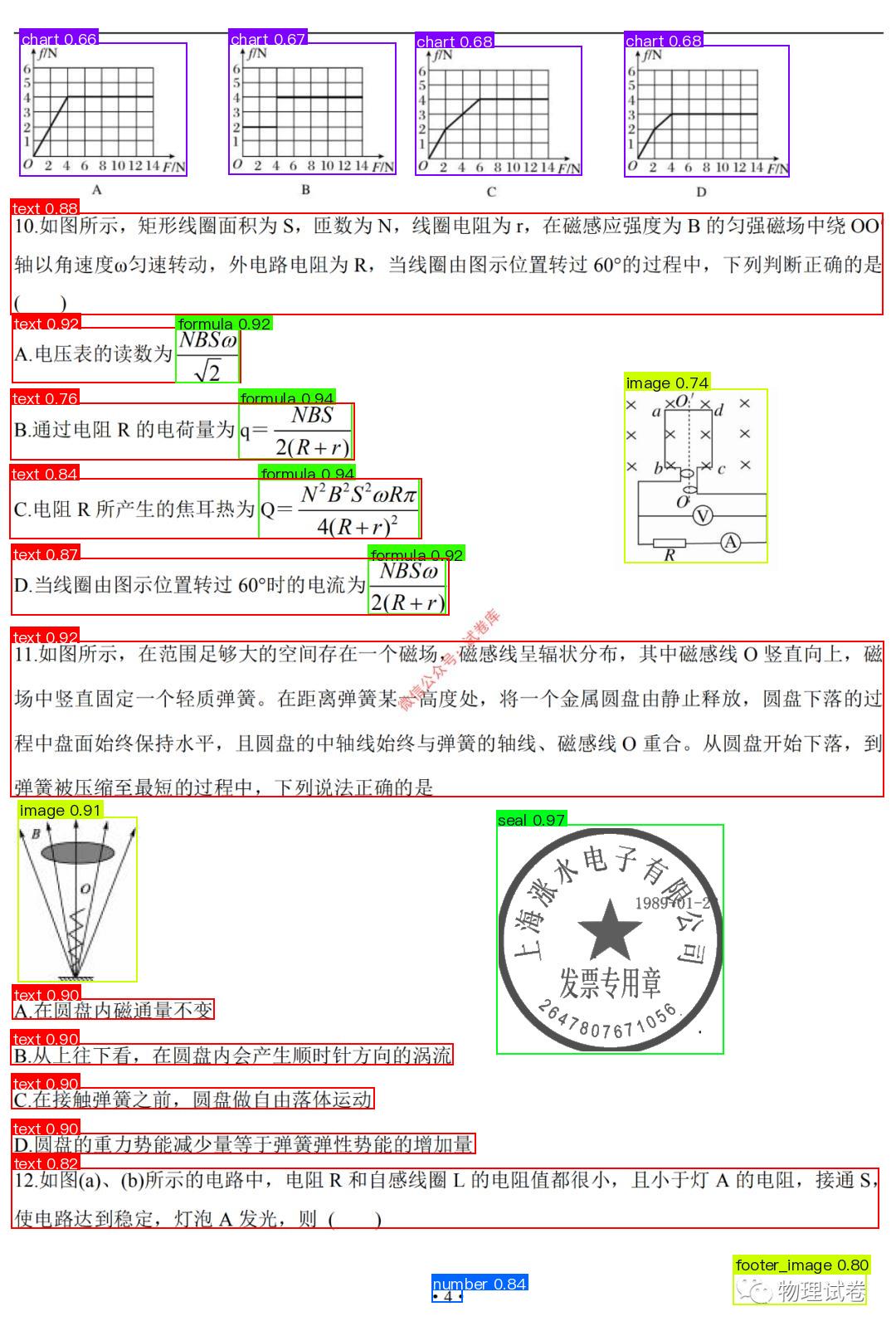} \\
\end{tabular}
\caption{Comparison of our results and DocLayout-YOLO \cite{zhao2024doclayout}.}
\label{tab:compare}
\end{figure*}

In this section, we compare the results of our proposed method with the advanced method DocLayout-YOLO \cite{zhao2024doclayout} through visualization. Due to differences in label categories, traditional quantitative metrics cannot be directly applied. Therefore, we employ visualization techniques to present the performance of each method enabling an intuitive comparison. 

The visualization results are shown in \textbf{Figure~\ref{tab:compare}}. The first row presents the results of DocLayout-YOLO \cite{zhao2024doclayout}, while the second row shows the results of our method. From the first column of images, it can be observed that our results include elements such as document titles, abstracts， paragraph titles and text, which are essential for understanding the semantic hierarchy and logical structure of the document. In contrast, DocLayout-YOLO categorizes these elements into only two broad classes, ``title'' and ``plain text,'' which limits its ability to parse the document's semantic layers effectively. Additionally, our PP-DocLayout can accurately locate page numbers, headers, footers, and footnotes, whereas DocLayout-YOLO often classifies these as ``abandon,'' disregarding their potential value.

The second column highlights the difference in formula recognition. Our method is capable of identifying both inline and block-level formulas, which is crucial for downstream tasks such as PDF-to-Markdown conversion. In contrast, DocLayout-YOLO struggles to recognize inline formulas, focusing only on prominent block-level formulas, which hinders its utility in tasks requiring full-text recognition.

In the third column, the comparison demonstrates our method's superior performance in handling handwritten notes. While our approach correctly identifies and categorizes handwritten content, DocLayout-YOLO misclassifies it as ``figure'' failing to capture its textual significance.

Finally, the fourth column illustrates our method's ability to distinguish between natural images, charts, and seals. Charts and seals are particularly important as they often contain high-value information, and our method ensures they are categorized separately. DocLayout-YOLO, on the other hand, does not make this distinction, potentially overlooking critical details.

Overall, our method provides a more granular and accurate representation of document elements, enabling better semantic understanding and supporting a wider range of downstream tasks compared to DocLayout-YOLO.

\subsection{Ablation Study}

In order to evaluate the impact of semi-supervised learning and knowledge distillation on model performance, we conducted a series of ablation experiments using the PP-DocLayout model variants. We compared the performance of each model with and without these techniques, measuring the mean Average Precision (mAP) at an IoU threshold of 0.5.

\textbf{Knowledge Distillation} We examined the effect of knowledge distillation on the PP-DocLayout-L variant as shown in \textbf{Table \ref{tab:ablation_study}}, The use of knowledge distillation yielded an mAP improvement from 89.3\% to 90.4\%, demonstrating a positive impact on model accuracy.

\begin{table}[h]
    \centering
    \resizebox{\columnwidth}{!}{%
    \begin{tabular}{lcc}
        \toprule
        \textbf{Algorithm Name} & \textbf{Knowledge Distillation} & \textbf{mAP@0.5 (\%)} \\
        \midrule
        PP-DocLayout-L & No & 89.3 \\
        PP-DocLayout-L & Yes & 90.4 (+1.1) \\
        \bottomrule
    \end{tabular}%
    }
    \caption{Results of the ablation study on the effect of knowledge distillation.}
    \label{tab:ablation_study}
\end{table}

\textbf{Semi-supervised Learning} As shown in \textbf{Table \ref{tab:ablation_semi_supervised}}, for the PP-DocLayout-M and PP-DocLayout-S models, incorporating semi-supervised learning significantly enhanced performance. Specifically, the mAP for PP-DocLayout-M improved from 73.8\% to 75.2\%, reflecting a 1.4\% increase. Similarly, PP-DocLayout-S saw an increase from 66.2\% to 70.9\%, marking a substantial improvement of 4.7\%.

\begin{table}[h]
    \centering
    \resizebox{\columnwidth}{!}{%
    \begin{tabular}{lcc}
        \toprule
        \textbf{Algorithm Name} & \textbf{Semi-Supervised Learning} & \textbf{mAP@0.5 (\%)} \\
        \midrule
        PP-DocLayout-M & No  & 73.8 \\
        PP-DocLayout-M & Yes & 75.2 (+1.4) \\
        PP-DocLayout-S & No  & 66.2 \\
        PP-DocLayout-S & Yes & 70.9 (+3.7) \\
        \bottomrule
    \end{tabular}%
    }
    \caption{Ablation study results on the effect of semi-supervised learning.}
    \label{tab:ablation_semi_supervised}
\end{table}

These results underscore the efficacy of both semi-supervised learning and knowledge distillation in enhancing model performance, thereby supporting their application in document layout analysis tasks.

\section{Conclusion}

We have introduced PP-DocLayout, a novel document layout detection model developed within the PaddlePaddle framework, to address the significant challenges faced by existing layout detection models in document intelligence. PP-DocLayout represents a significant step forward in document layout analysis, offering a versatile and efficient solution to address the complexities and diversities of document structures. Our models not only push the boundaries of the state of the art but also provide practical tools for real-world applications, paving the way for future advancements in document intelligence and related fields.


{
    \small
    \bibliographystyle{ieeenat_fullname}
    \bibliography{main}

\begin{thebibliography}{15}
\providecommand{\natexlab}[1]{#1}
\providecommand{\url}[1]{\texttt{#1}}
\expandafter\ifx\csname urlstyle\endcsname\relax
  \providecommand{\doi}[1]{doi: #1}\else
  \providecommand{\doi}{doi: \begingroup \urlstyle{rm}\Url}\fi

\bibitem[Da et~al.(2023)Da, Luo, Zheng, and Yao]{da2023vision}
Cheng Da, Chuwei Luo, Qi Zheng, and Cong Yao.
\newblock Vision grid transformer for document layout analysis.
\newblock In \emph{Proceedings of the IEEE/CVF international conference on computer vision}, pages 19462--19472, 2023.

\bibitem[Hinton et~al.(2015)Hinton, Vinyals, and Dean]{hinton2015distilling}
Geoffrey Hinton, Oriol Vinyals, and Jeff Dean.
\newblock Distilling the knowledge in a neural network.
\newblock \emph{arXiv preprint arXiv:1503.02531}, 2015.

\bibitem[Huang et~al.(2022)Huang, Lv, Cui, Lu, and Wei]{huang2022layoutlmv3}
Yupan Huang, Tengchao Lv, Lei Cui, Yutong Lu, and Furu Wei.
\newblock Layoutlmv3: Pre-training for document ai with unified text and image masking.
\newblock In \emph{Proceedings of the 30th ACM International Conference on Multimedia}, pages 4083--4091, 2022.

\bibitem[Li et~al.(2022)Li, Xu, Lv, Cui, Zhang, and Wei]{li2022dit}
Junlong Li, Yiheng Xu, Tengchao Lv, Lei Cui, Cha Zhang, and Furu Wei.
\newblock Dit: Self-supervised pre-training for document image transformer.
\newblock In \emph{Proceedings of the 30th ACM International Conference on Multimedia}, pages 3530--3539, 2022.

\bibitem[Loshchilov and Hutter(2016)]{loshchilov2016sgdr}
Ilya Loshchilov and Frank Hutter.
\newblock Sgdr: Stochastic gradient descent with warm restarts.
\newblock \emph{arXiv preprint arXiv:1608.03983}, 2016.

\bibitem[Pfitzmann et~al.(2022)Pfitzmann, Auer, Dolfi, Nassar, and Staar]{pfitzmann2022doclaynet}
Birgit Pfitzmann, Christoph Auer, Michele Dolfi, Ahmed~S Nassar, and Peter Staar.
\newblock Doclaynet: A large human-annotated dataset for document-layout segmentation.
\newblock In \emph{Proceedings of the 28th ACM SIGKDD conference on knowledge discovery and data mining}, pages 3743--3751, 2022.

\bibitem[Ren et~al.(2016)Ren, He, Girshick, and Sun]{ren2016faster}
Shaoqing Ren, Kaiming He, Ross Girshick, and Jian Sun.
\newblock Faster r-cnn: Towards real-time object detection with region proposal networks.
\newblock \emph{IEEE transactions on pattern analysis and machine intelligence}, 39\penalty0 (6):\penalty0 1137--1149, 2016.

\bibitem[Wang et~al.(2024)Wang, Chen, Liu, Chen, Lin, Han, and Ding]{wang2024yolov10}
Ao Wang, Hui Chen, Lihao Liu, Kai Chen, Zijia Lin, Jungong Han, and Guiguang Ding.
\newblock Yolov10: Real-time end-to-end object detection.
\newblock \emph{arXiv preprint arXiv:2405.14458}, 2024.

\bibitem[Wei et~al.(2024)Wei, Liu, Chen, Wang, Kong, Xu, Ge, Zhao, Sun, Peng, et~al.]{wei2024general}
Haoran Wei, Chenglong Liu, Jinyue Chen, Jia Wang, Lingyu Kong, Yanming Xu, Zheng Ge, Liang Zhao, Jianjian Sun, Yuang Peng, et~al.
\newblock General ocr theory: Towards ocr-2.0 via a unified end-to-end model.
\newblock 2024.

\bibitem[Xu et~al.(2020{\natexlab{a}})Xu, Li, Cui, Huang, Wei, and Zhou]{xu2020layoutlm}
Yiheng Xu, Minghao Li, Lei Cui, Shaohan Huang, Furu Wei, and Ming Zhou.
\newblock Layoutlm: Pre-training of text and layout for document image understanding.
\newblock In \emph{Proceedings of the 26th ACM SIGKDD international conference on knowledge discovery \& data mining}, pages 1192--1200, 2020{\natexlab{a}}.

\bibitem[Xu et~al.(2020{\natexlab{b}})Xu, Xu, Lv, Cui, Wei, Wang, Lu, Florencio, Zhang, Che, et~al.]{xu2020layoutlmv2}
Yang Xu, Yiheng Xu, Tengchao Lv, Lei Cui, Furu Wei, Guoxin Wang, Yijuan Lu, Dinei Florencio, Cha Zhang, Wanxiang Che, et~al.
\newblock Layoutlmv2: Multi-modal pre-training for visually-rich document understanding.
\newblock \emph{arXiv preprint arXiv:2012.14740}, 2020{\natexlab{b}}.

\bibitem[Yu et~al.(2021)Yu, Chang, Lv, Xu, Cui, Ji, Dang, Deng, Wang, Du, et~al.]{yu2021pp}
Guanghua Yu, Qinyao Chang, Wenyu Lv, Chang Xu, Cheng Cui, Wei Ji, Qingqing Dang, Kaipeng Deng, Guanzhong Wang, Yuning Du, et~al.
\newblock Pp-picodet: A better real-time object detector on mobile devices.
\newblock \emph{arXiv preprint arXiv:2111.00902}, 2021.

\bibitem[Zhao et~al.(2024{\natexlab{a}})Zhao, Lv, Xu, Wei, Wang, Dang, Liu, and Chen]{zhao2024detrs}
Yian Zhao, Wenyu Lv, Shangliang Xu, Jinman Wei, Guanzhong Wang, Qingqing Dang, Yi Liu, and Jie Chen.
\newblock Detrs beat yolos on real-time object detection.
\newblock In \emph{Proceedings of the IEEE/CVF Conference on Computer Vision and Pattern Recognition}, pages 16965--16974, 2024{\natexlab{a}}.

\bibitem[Zhao et~al.(2024{\natexlab{b}})Zhao, Kang, Wang, and He]{zhao2024doclayout}
Zhiyuan Zhao, Hengrui Kang, Bin Wang, and Conghui He.
\newblock Doclayout-yolo: Enhancing document layout analysis through diverse synthetic data and global-to-local adaptive perception.
\newblock \emph{arXiv preprint arXiv:2410.12628}, 2024{\natexlab{b}}.

\bibitem[Zhong et~al.(2019)Zhong, Tang, and Yepes]{zhong2019publaynet}
Xu Zhong, Jianbin Tang, and Antonio~Jimeno Yepes.
\newblock Publaynet: largest dataset ever for document layout analysis.
\newblock In \emph{2019 International conference on document analysis and recognition (ICDAR)}, pages 1015--1022. IEEE, 2019.

\end{thebibliography}
}

\clearpage 
\appendix
\section{Appendix}

This appendix provides additional visualization results and datset details that support the main content of the paper.

\subsection{Instance counts}

\textbf{Table~\ref{tab:instance_counts}} provides a detailed breakdown of the instance counts for each category in the training and validation datasets. The dataset covers a wide range of document elements, including paragraph titles, images, text, formulas, tables, and more. This distribution underscores the diversity and complexity of the dataset, which is essential for developing robust document layout analysis models.

\begin{table}[h]
    \centering
    \small 
    \resizebox{\columnwidth}{!}{%
    \begin{tabular}{lccc}
        \toprule
        \textbf{Category} & \textbf{ID} & \textbf{Training} & \textbf{Validation} \\
        \textbf{} & \textbf{} & \textbf{Instances} & \textbf{Instances} \\
        \midrule
        paragraph\_title & 0 & 42158 & 715 \\
        image           & 1 & 11455 & 230 \\
        text            & 2 & 217257 & 3342 \\
        number          & 3 & 25217 & 430 \\
        abstract        & 4 & 2067 & 15 \\
        content         & 5 & 643 & 7 \\
        figure\_title   & 6 & 8136 & 161 \\
        formula         & 7 & 113145 & 1961 \\
        table           & 8 & 6258 & 127 \\
        table\_title    & 9 & 5553 & 109 \\
        reference       & 10 & 2217 & 32 \\
        doc\_title      & 11 & 4276 & 35 \\
        footnote        & 12 & 3566 & 38 \\
        header          & 13 & 25001 & 430 \\
        algorithm       & 14 & 289 & 10 \\
        footer          & 15 & 11546 & 245 \\
        seal            & 16 & 3144 & 52 \\
        chart\_title    & 17 & 11138 & 247 \\
        chart           & 18 & 10849 & 303 \\
        formula\_number & 19 & 15668 & 359 \\
        header\_image   & 20 & 6599 & 171 \\
        footer\_image   & 21 & 1654 & 27 \\
        aside\_text     & 22 & 2093 & 32 \\
        \bottomrule
    \end{tabular}%
    }
    \caption{Instance counts for each category in the training and validation datasets.}
    \label{tab:instance_counts}
\end{table}

\subsection{Additional Visualization Results}

\begin{figure*}[htbp]
    \centering
    \begin{subfigure}[b]{\textwidth}
        \centering
        \begin{minipage}{0.19\textwidth}
            \includegraphics[width=\linewidth]{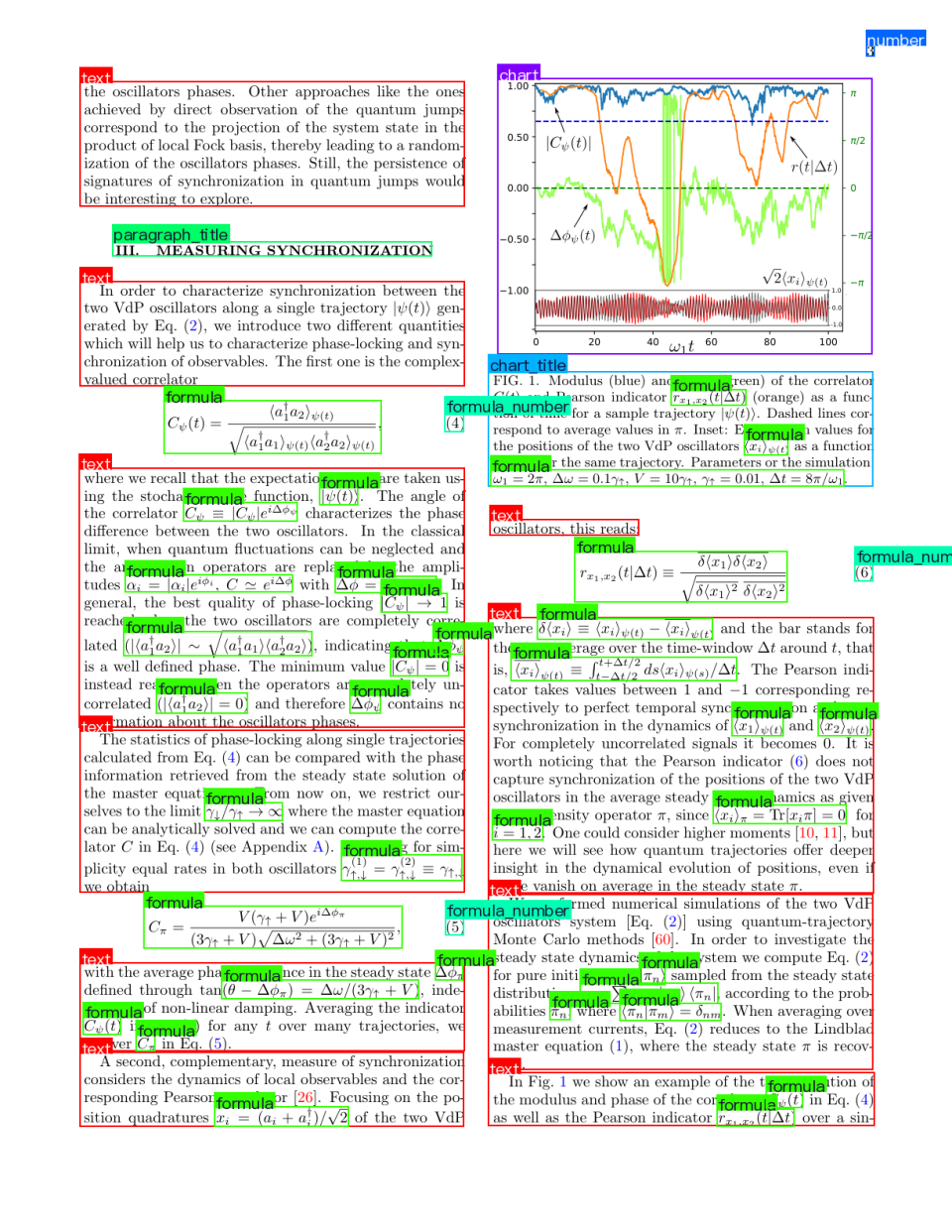}
        \end{minipage}
        \begin{minipage}{0.19\textwidth}
            \includegraphics[width=\linewidth]{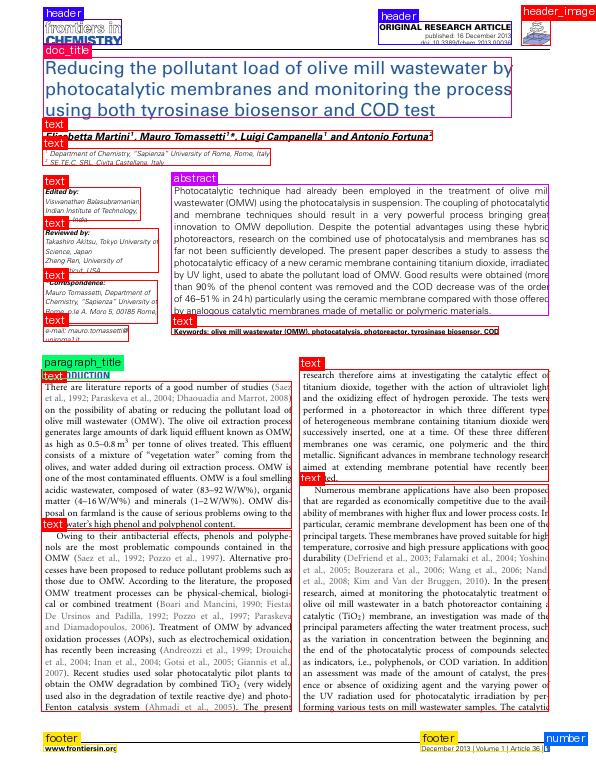}
        \end{minipage}
        \begin{minipage}{0.19\textwidth}
            \includegraphics[width=\linewidth]{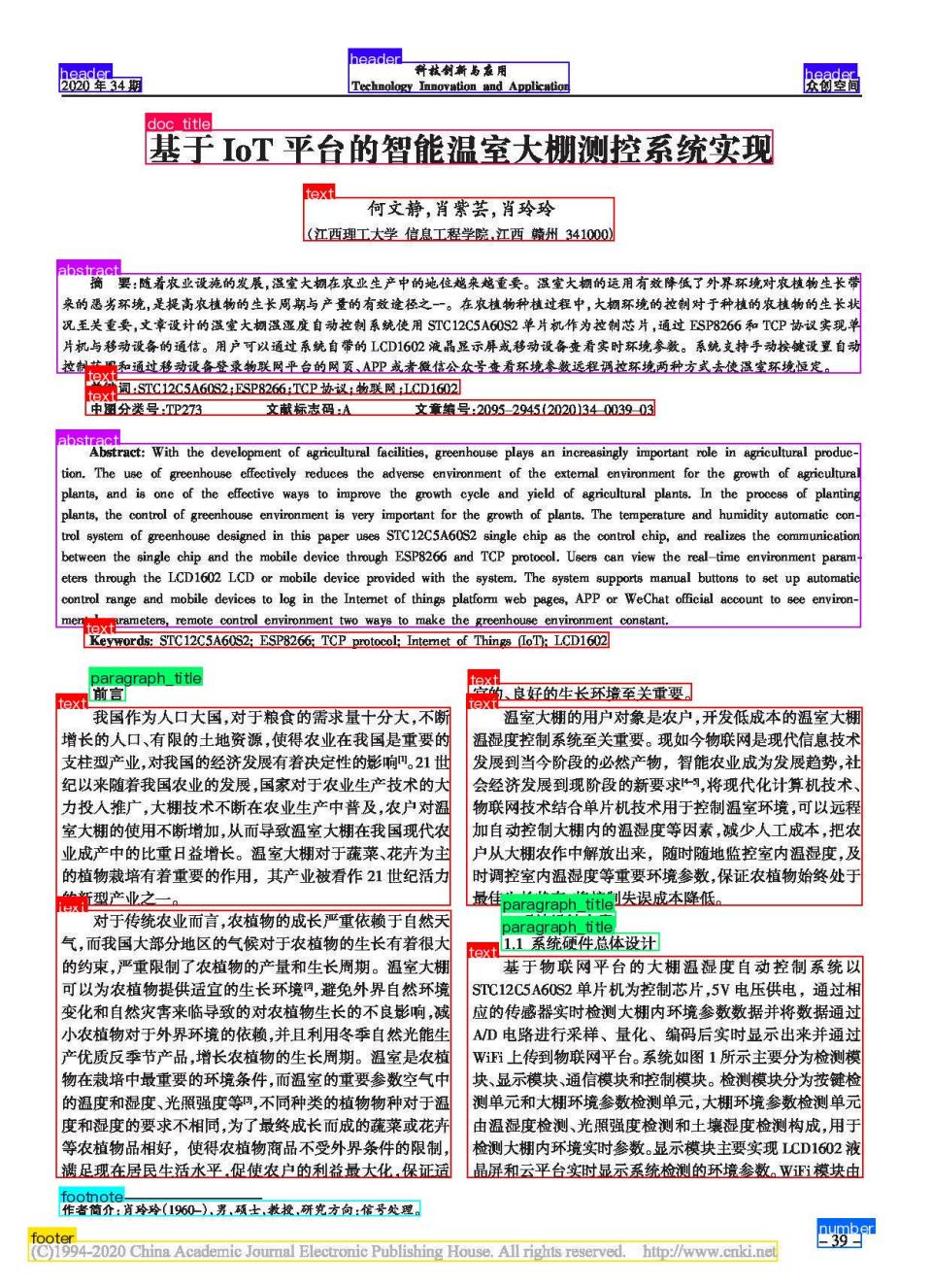}
        \end{minipage}
        \begin{minipage}{0.19\textwidth}
            \includegraphics[width=\linewidth]{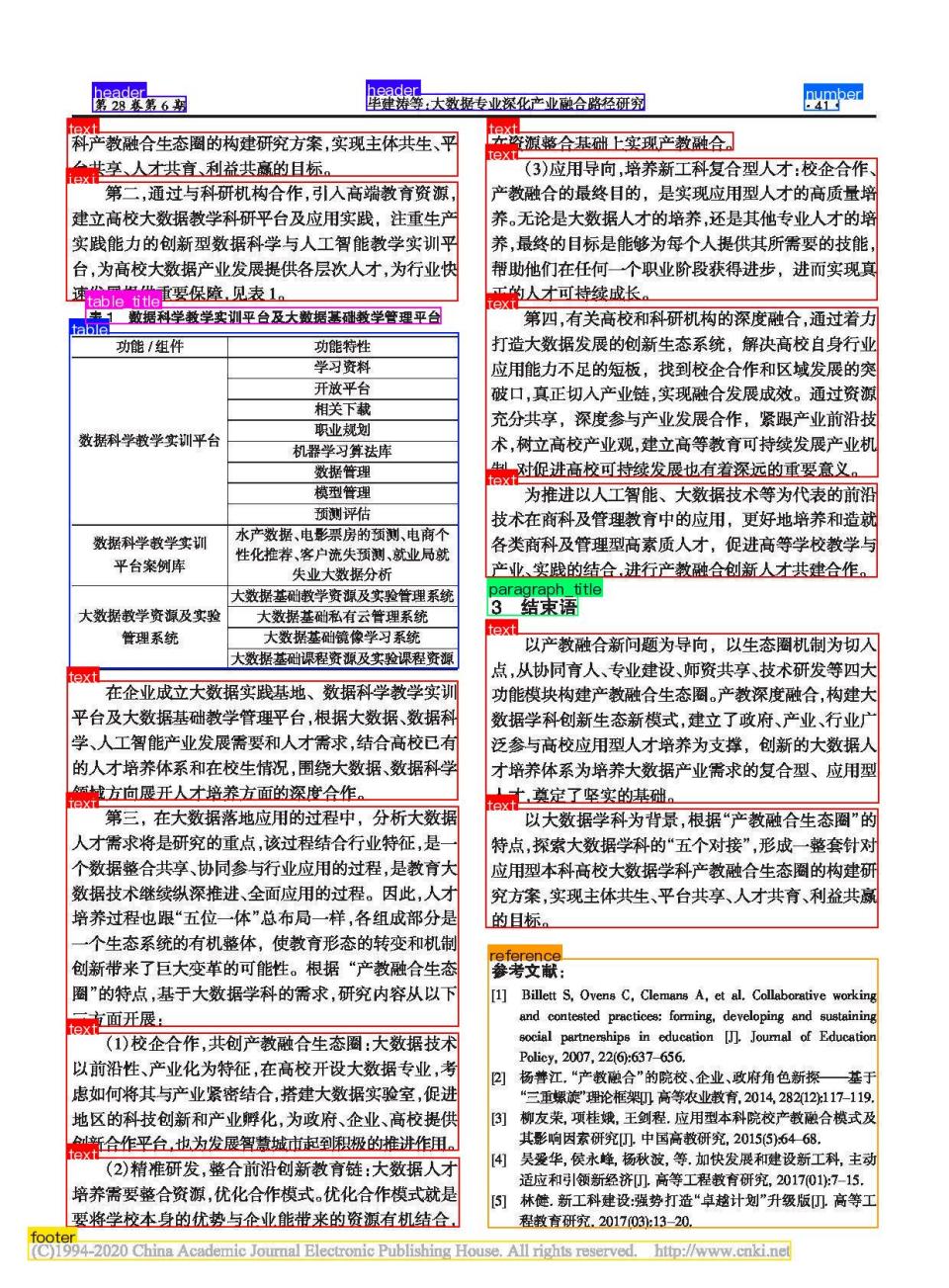}
        \end{minipage}
        \begin{minipage}{0.19\textwidth}
            \includegraphics[width=\linewidth]{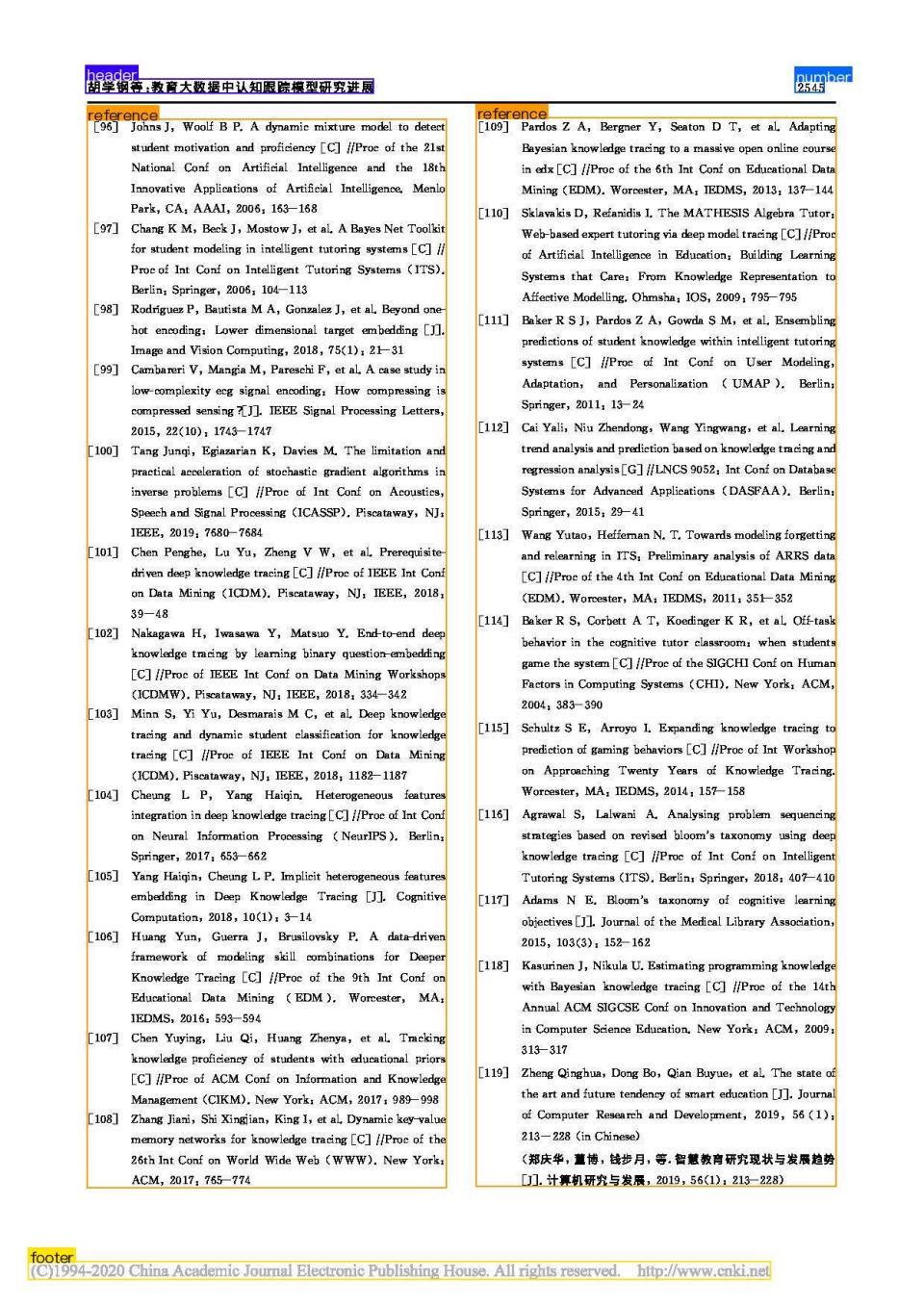}
        \end{minipage}
        \caption{Paper Visualization}
    \end{subfigure}
    
    \begin{subfigure}[b]{\textwidth}
        \centering
        \begin{minipage}{0.19\textwidth}
            \includegraphics[width=\linewidth]{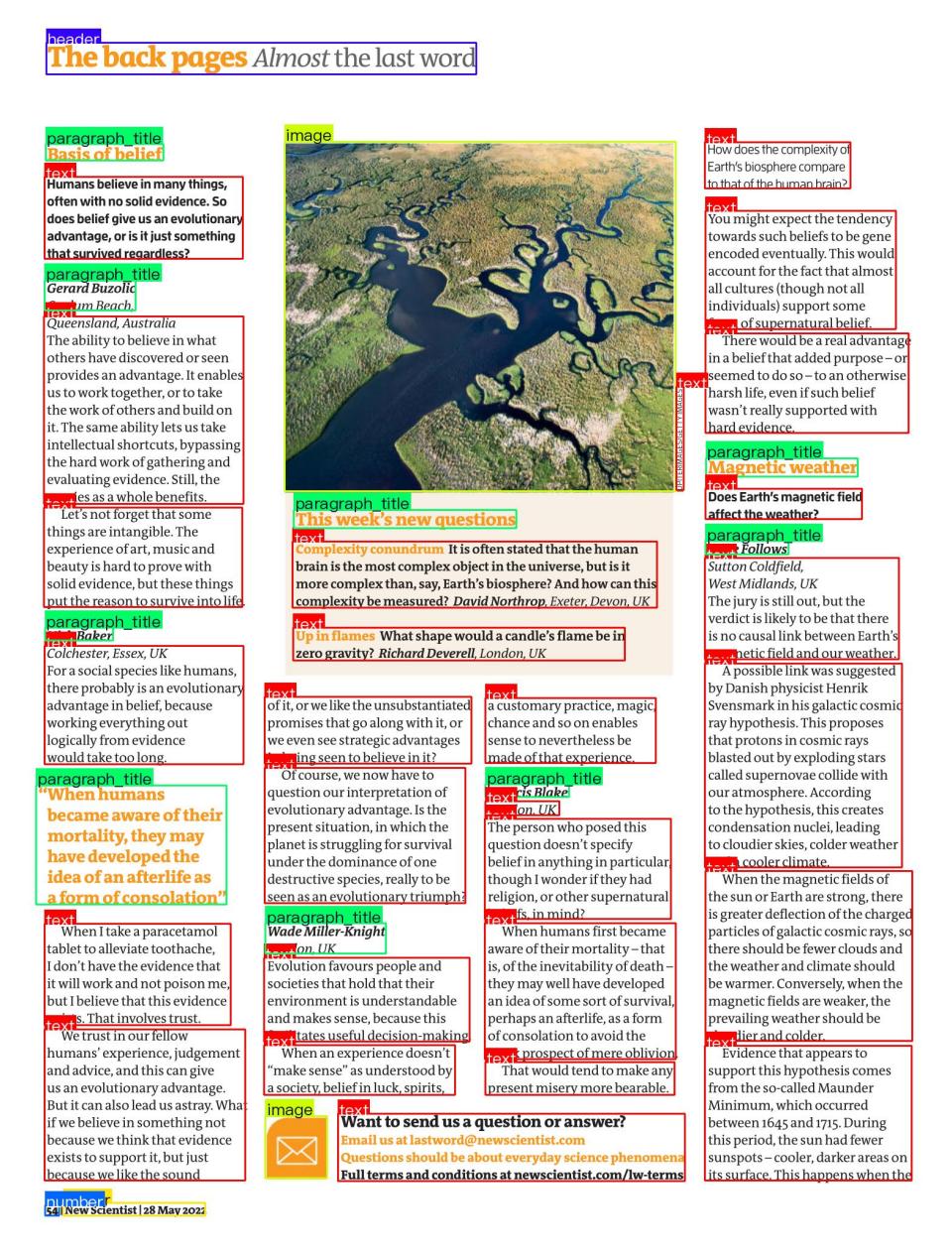}
        \end{minipage}
        \begin{minipage}{0.19\textwidth}
            \includegraphics[width=\linewidth]{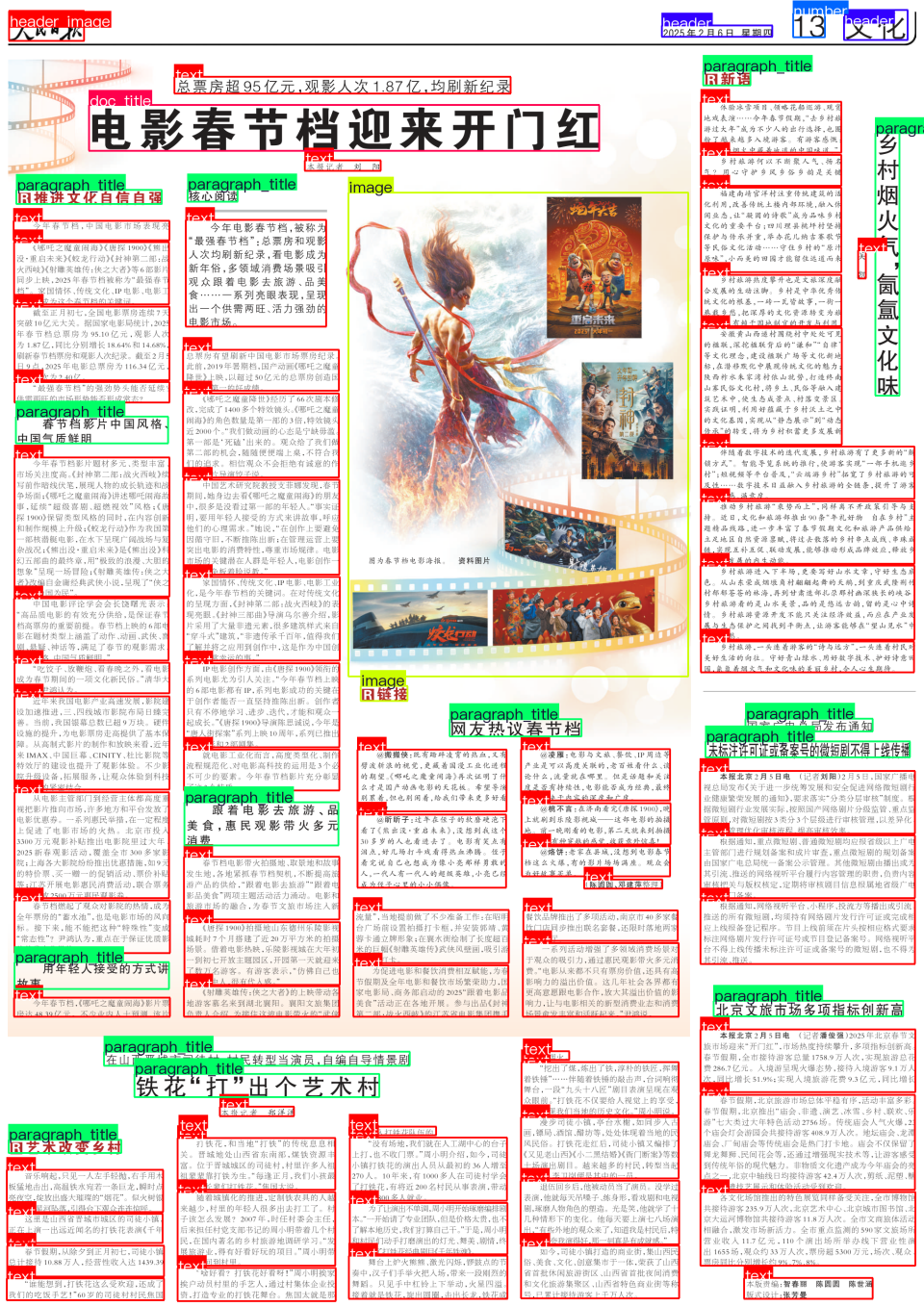}
        \end{minipage}
        \begin{minipage}{0.19\textwidth}
            \includegraphics[width=\linewidth]{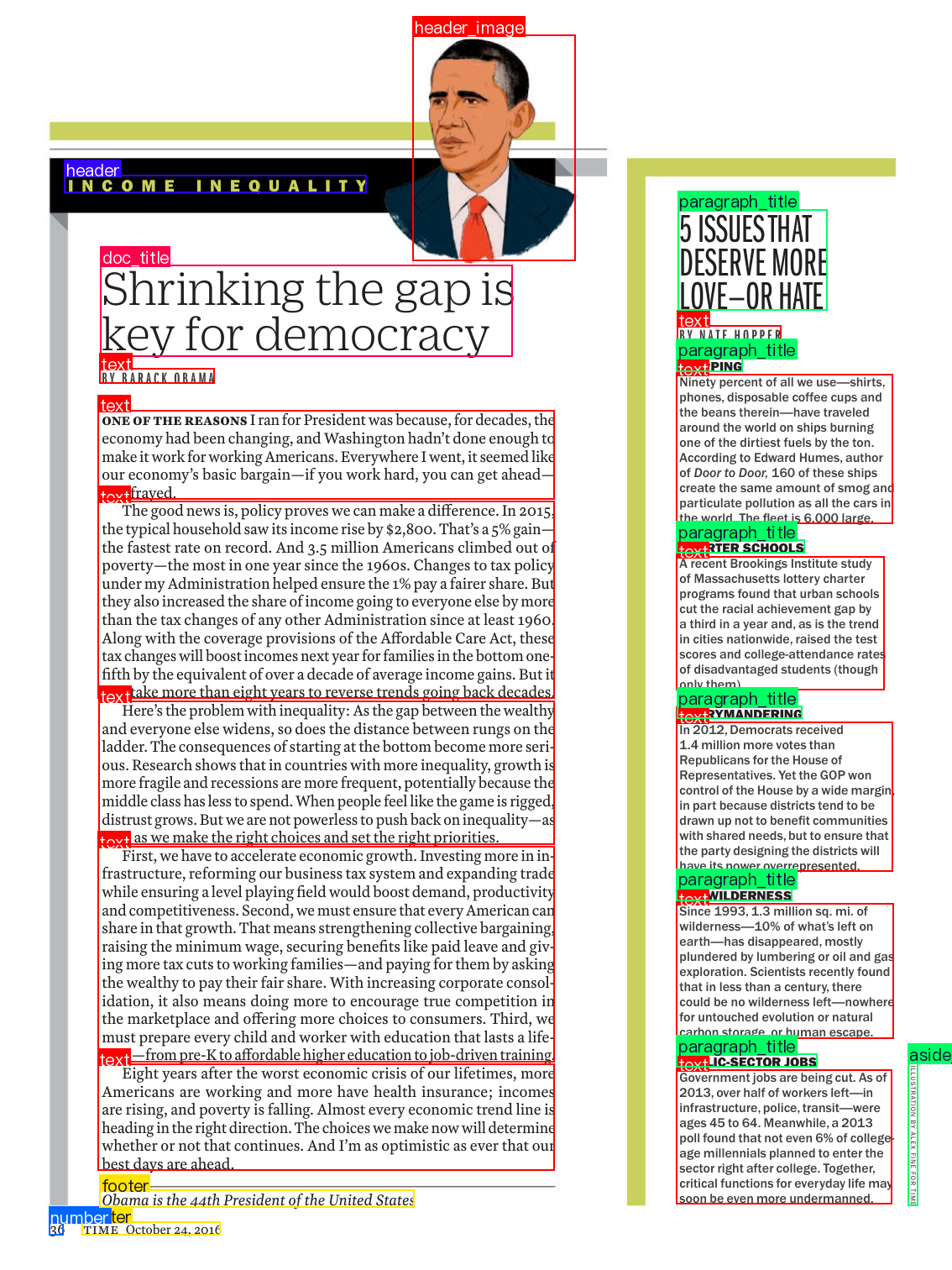}
        \end{minipage}
        \begin{minipage}{0.19\textwidth}
            \includegraphics[width=\linewidth]{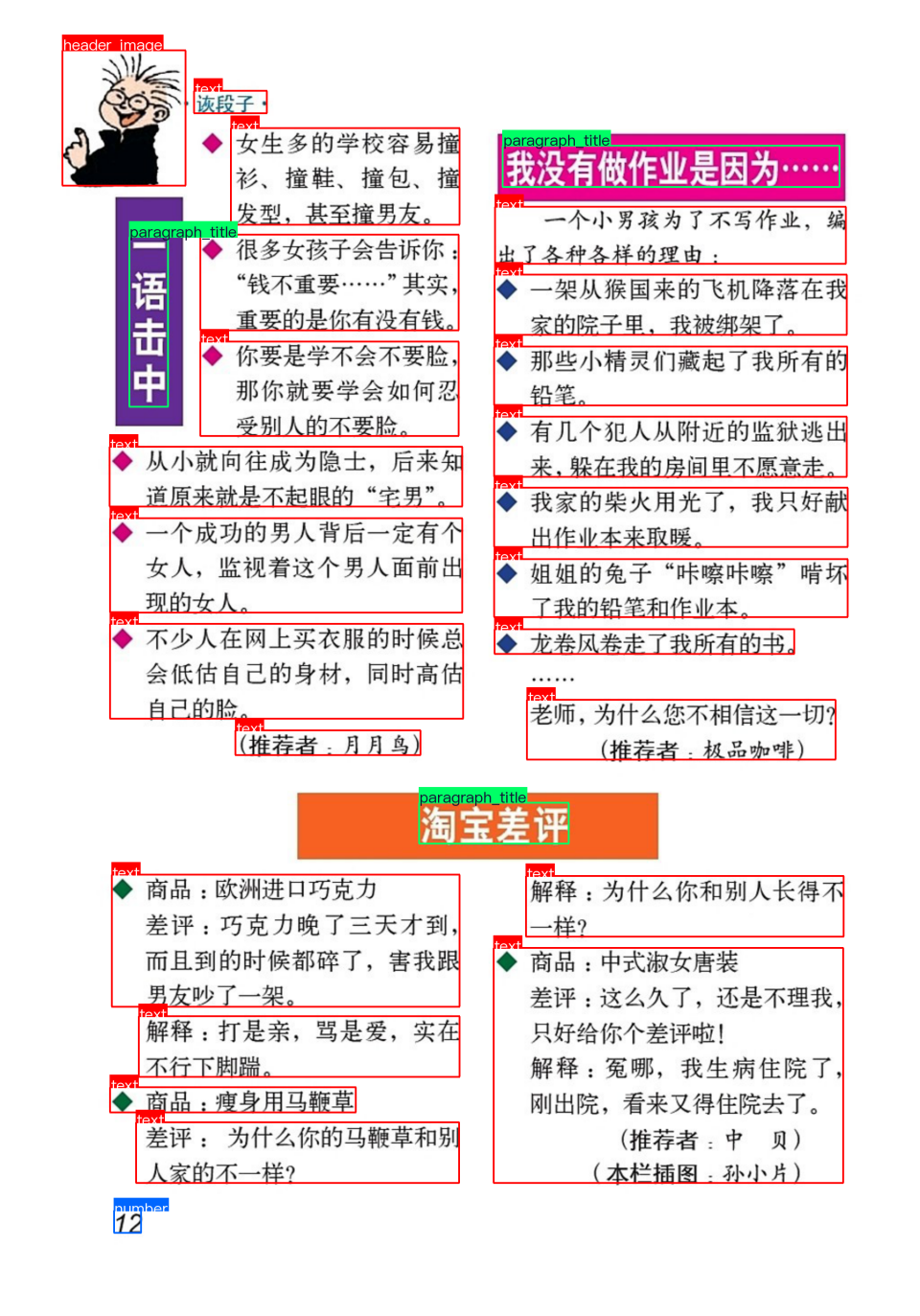}
        \end{minipage}
        \begin{minipage}{0.19\textwidth}
            \includegraphics[width=\linewidth]{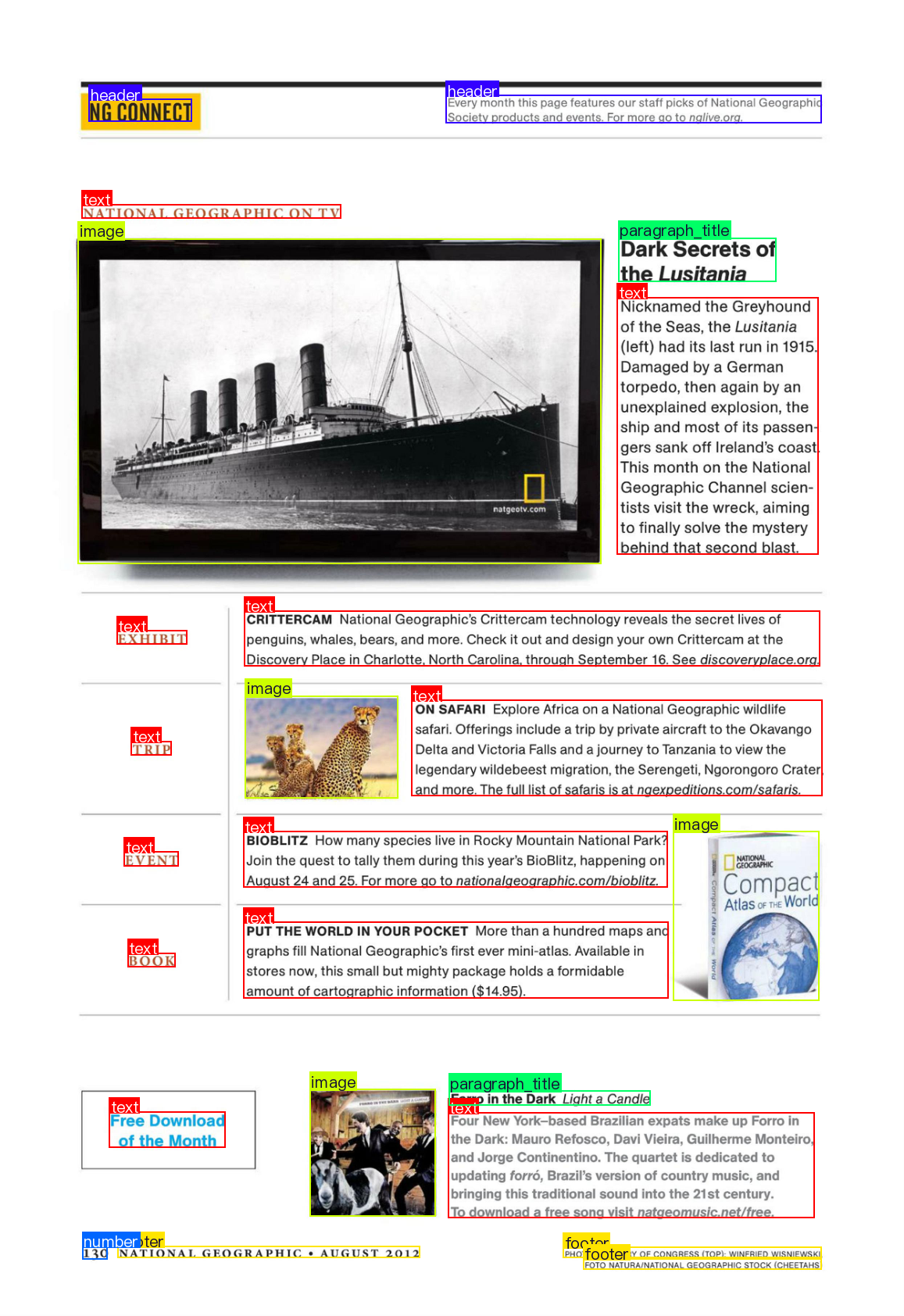}
        \end{minipage}
        \caption{Magazine and Newspaper Visualization}
    \end{subfigure}
    
    \begin{subfigure}[b]{\textwidth}
        \centering
        \begin{minipage}{0.19\textwidth}
            \includegraphics[width=\linewidth]{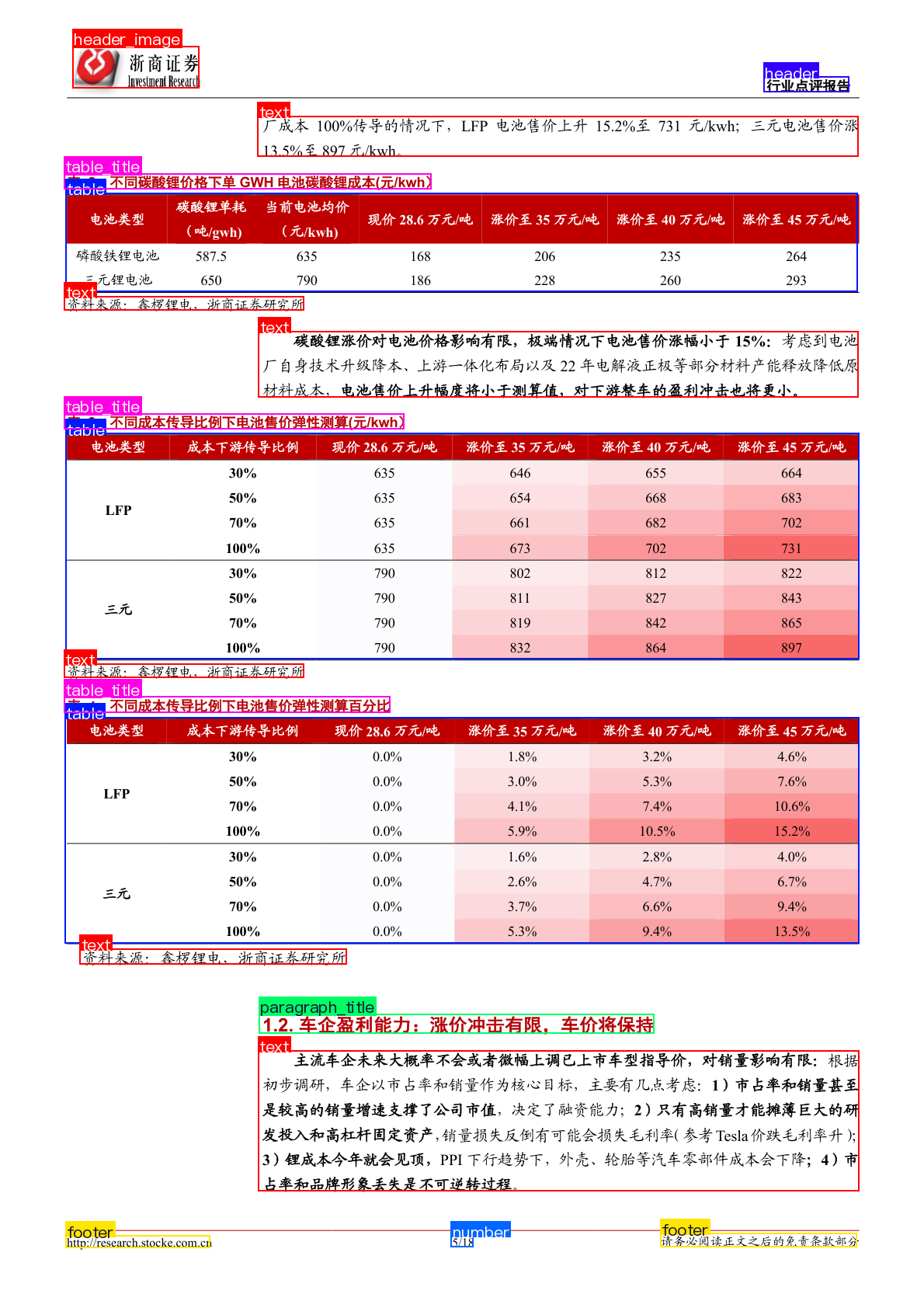}
        \end{minipage}
        \begin{minipage}{0.19\textwidth}
            \includegraphics[width=\linewidth]{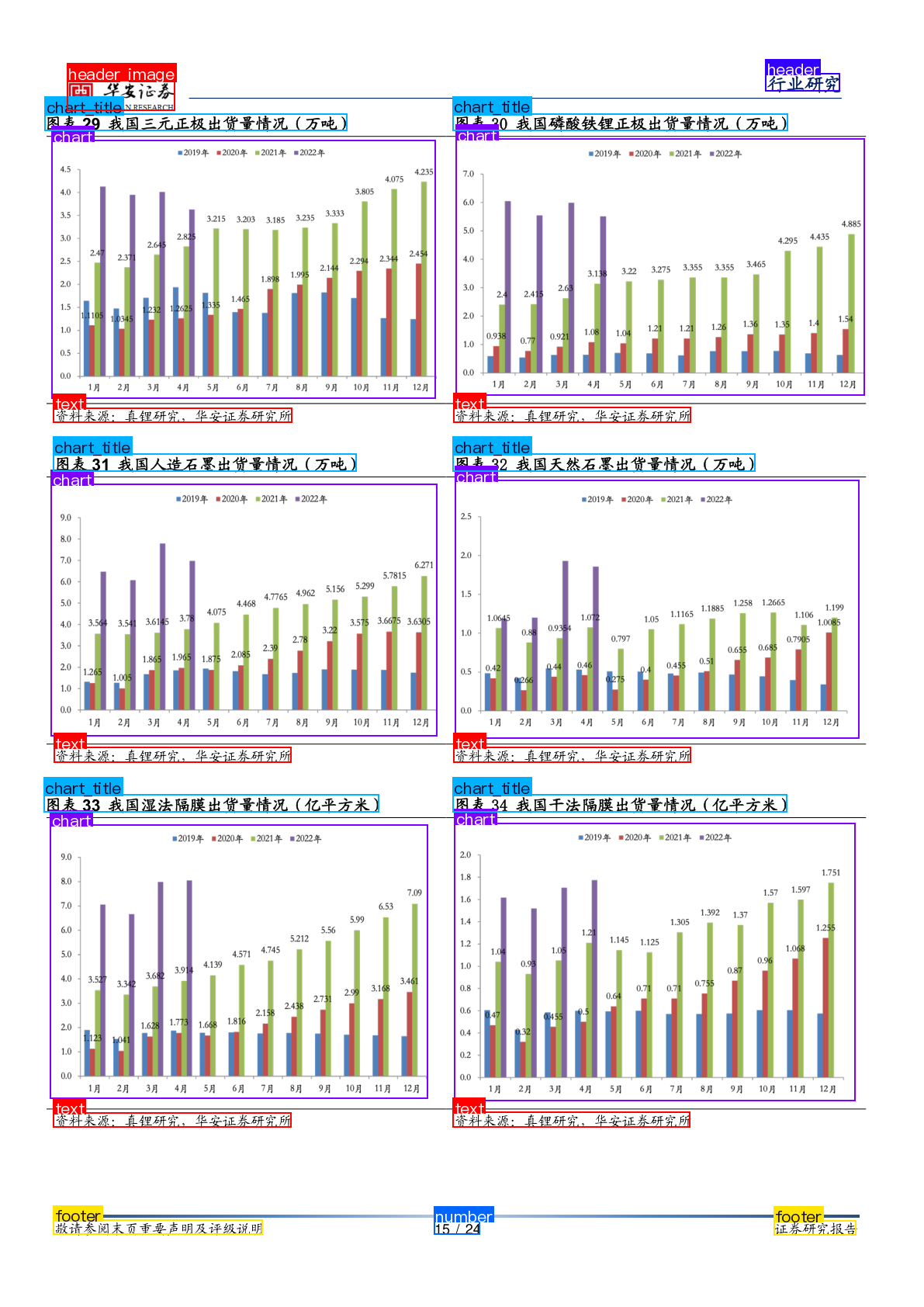}
        \end{minipage}
        \begin{minipage}{0.19\textwidth}
            \includegraphics[width=\linewidth]{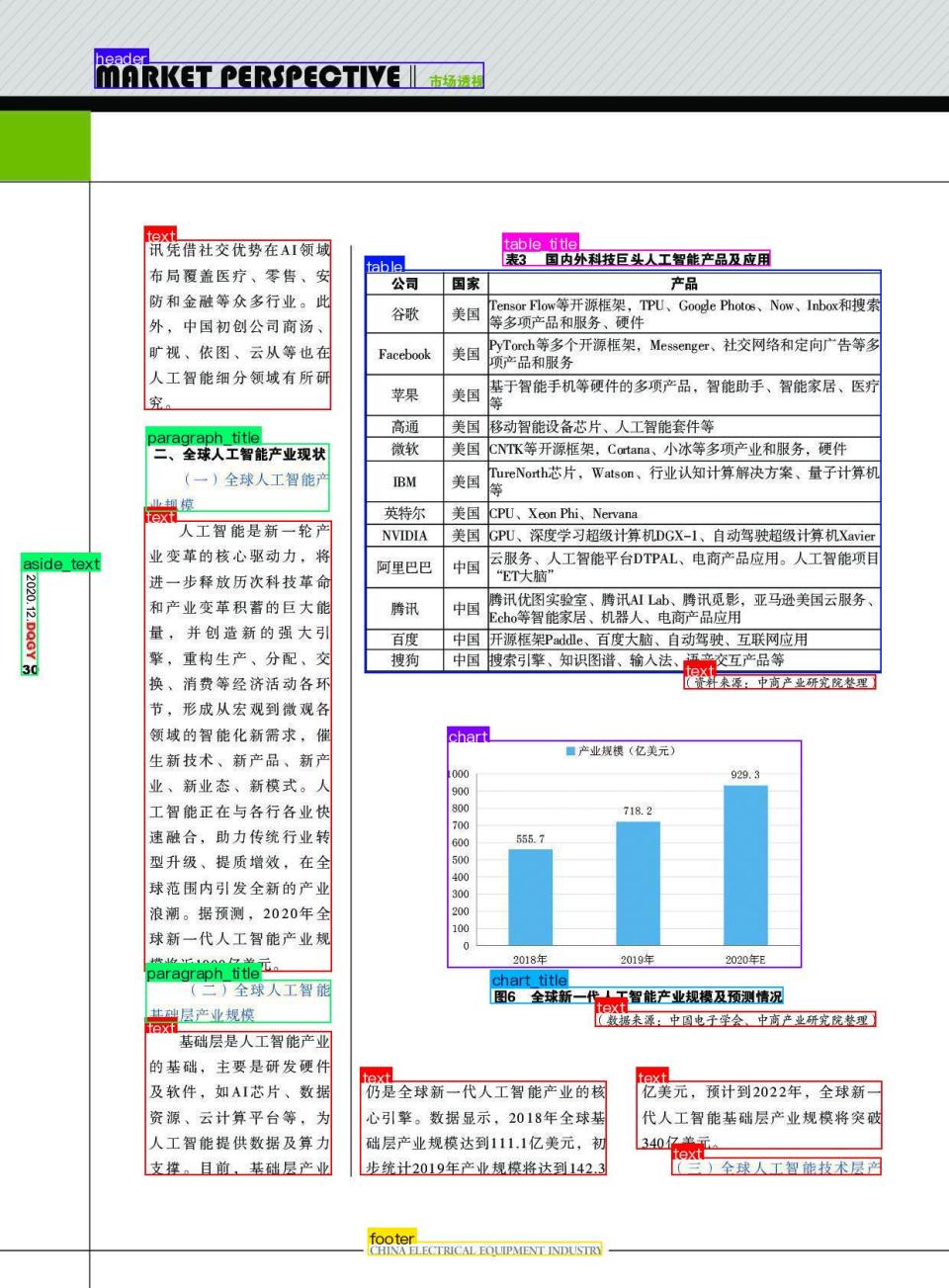}
        \end{minipage}
        \begin{minipage}{0.19\textwidth}
            \includegraphics[width=\linewidth]{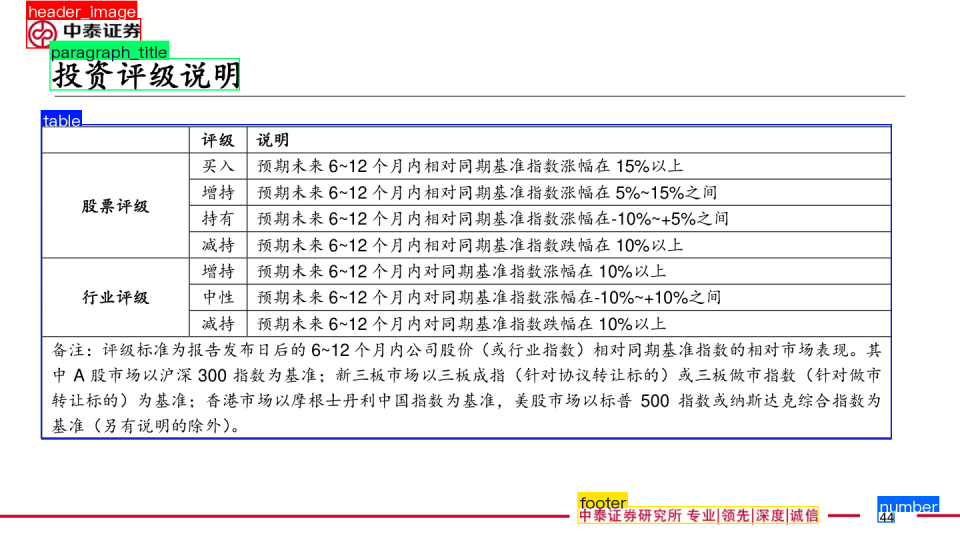}
        \end{minipage}
        \begin{minipage}{0.19\textwidth}
            \includegraphics[width=\linewidth]{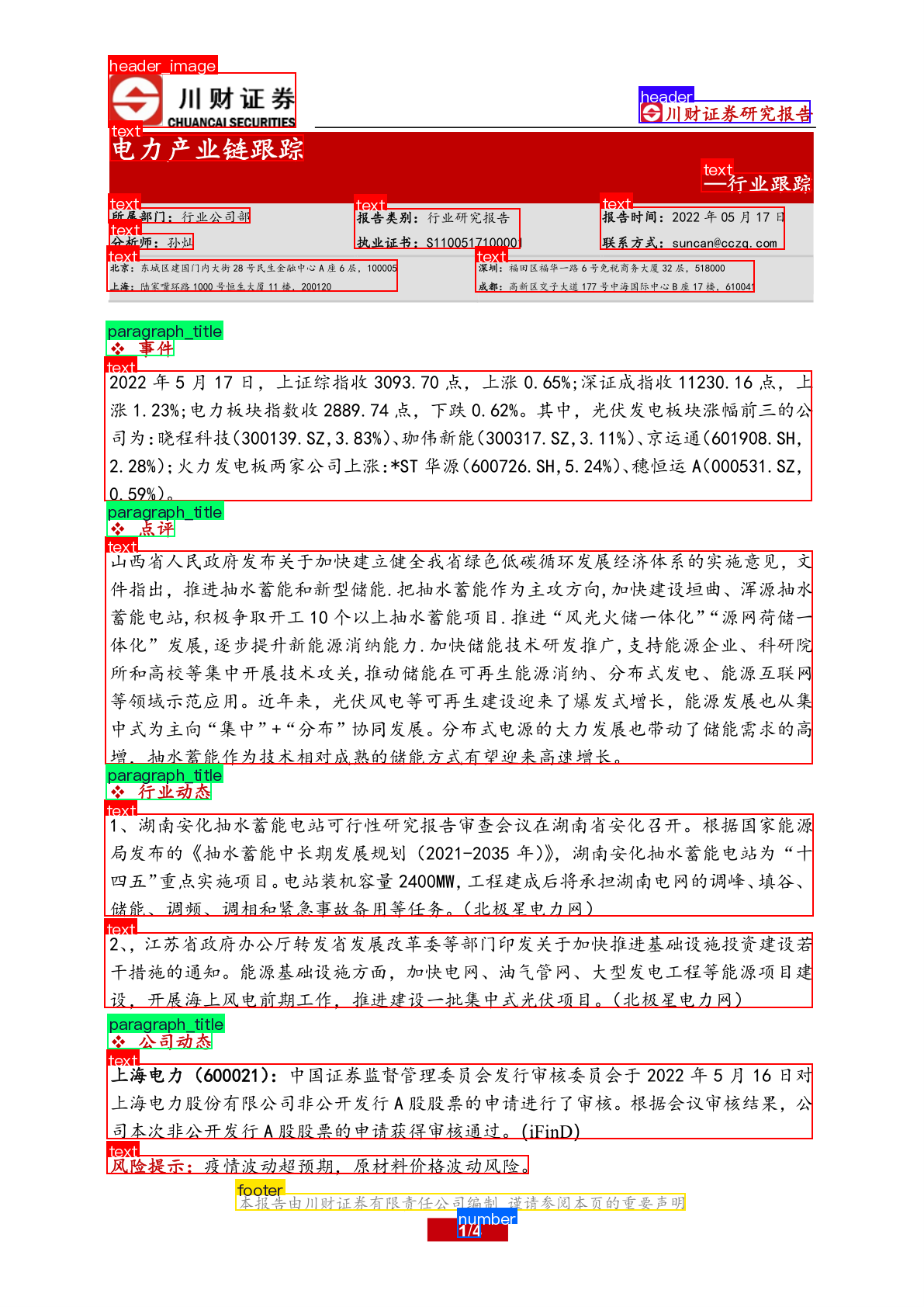}
        \end{minipage}
        \caption{Research Report Visualization}
    \end{subfigure}

    \begin{subfigure}[b]{\textwidth}
        \centering
        \begin{minipage}{0.19\textwidth}
            \includegraphics[width=\linewidth]{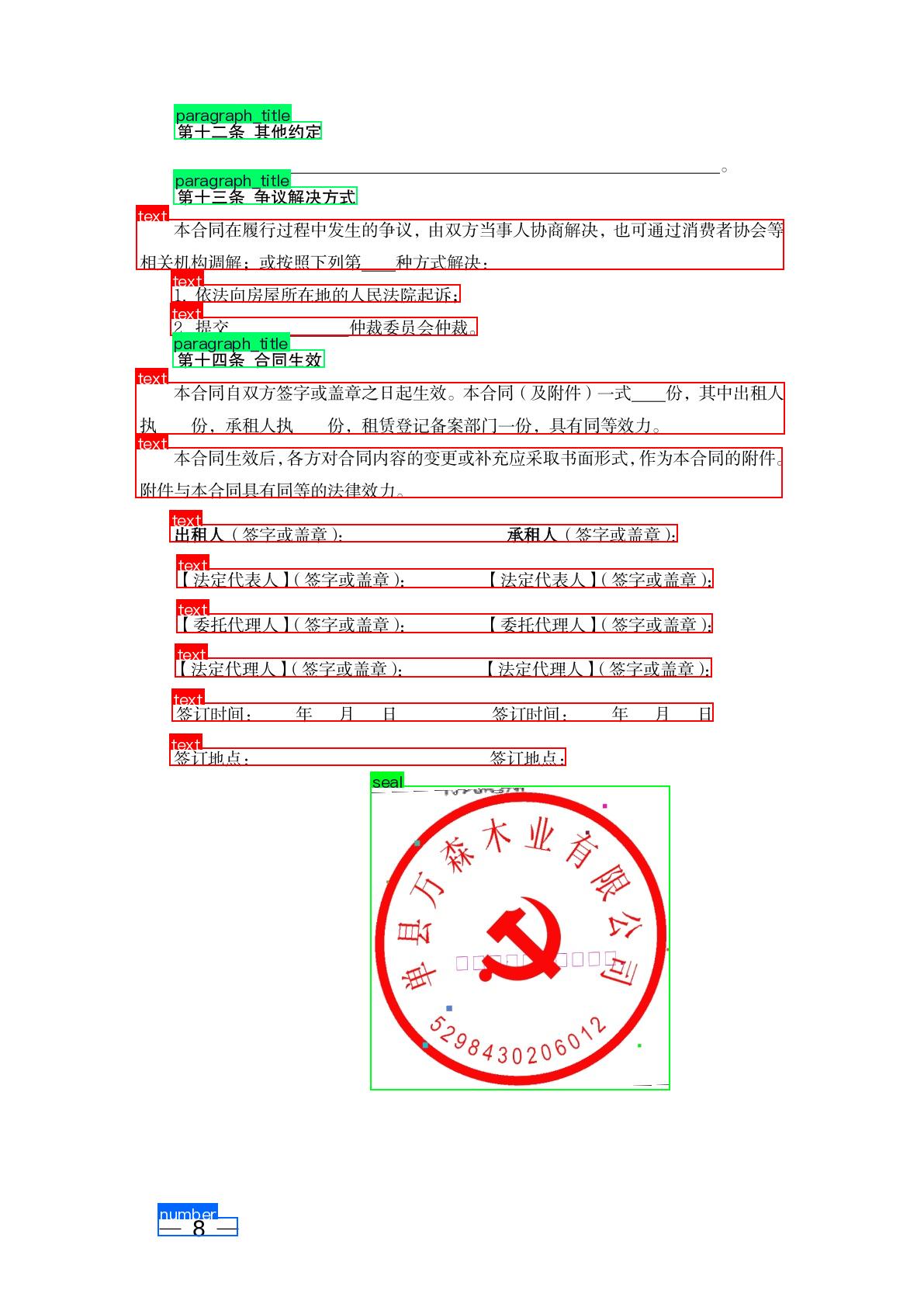}
        \end{minipage}
        \begin{minipage}{0.19\textwidth}
            \includegraphics[width=\linewidth]{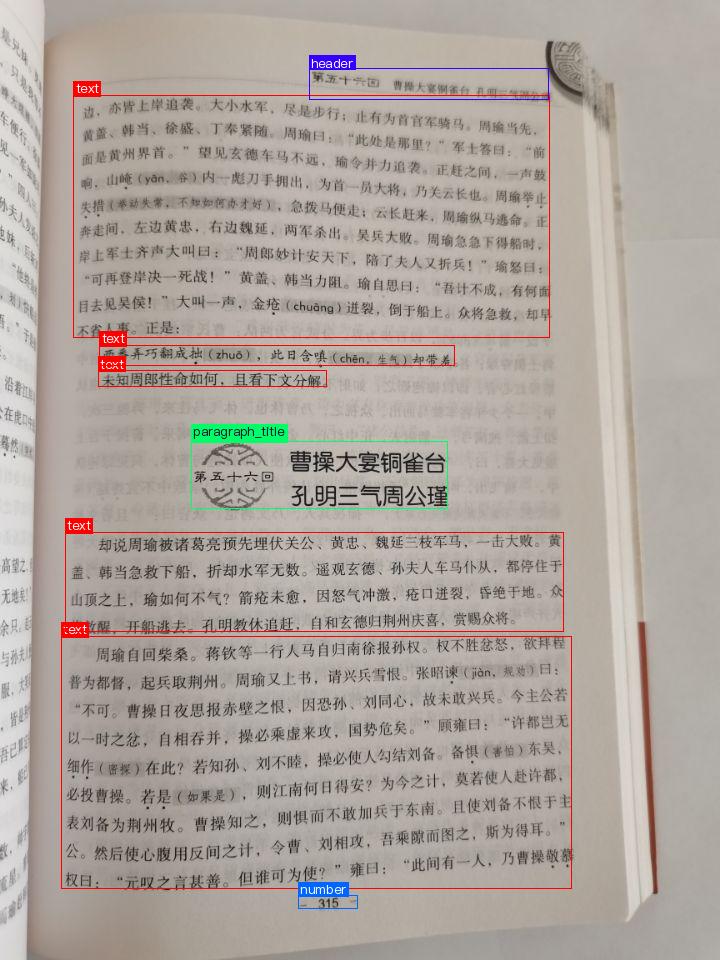}
        \end{minipage}
        \begin{minipage}{0.19\textwidth}
            \includegraphics[width=\linewidth]{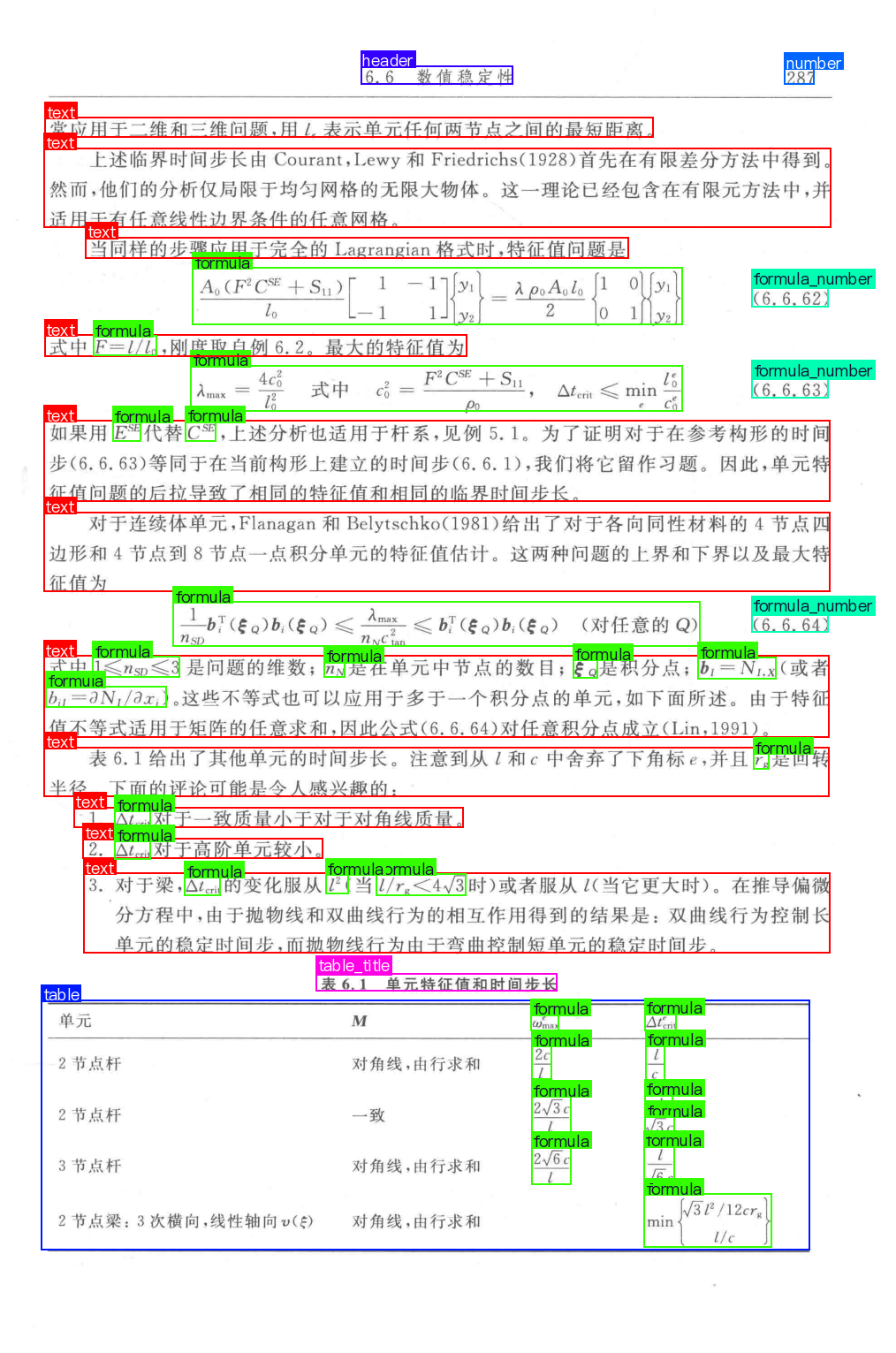}
        \end{minipage}
        \begin{minipage}{0.19\textwidth}
            \includegraphics[width=\linewidth]{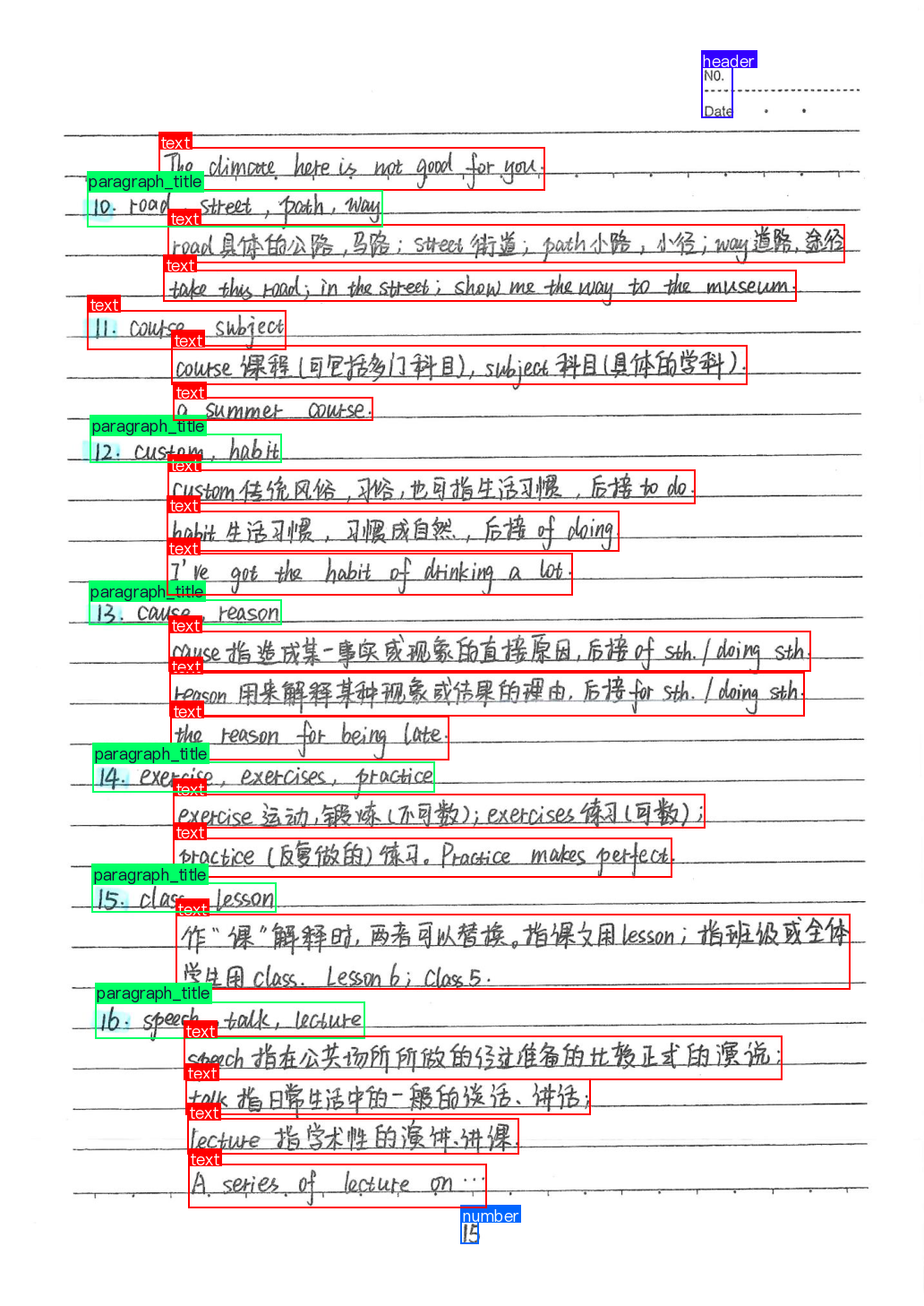}
        \end{minipage}
        \begin{minipage}{0.19\textwidth}
            \includegraphics[width=\linewidth]{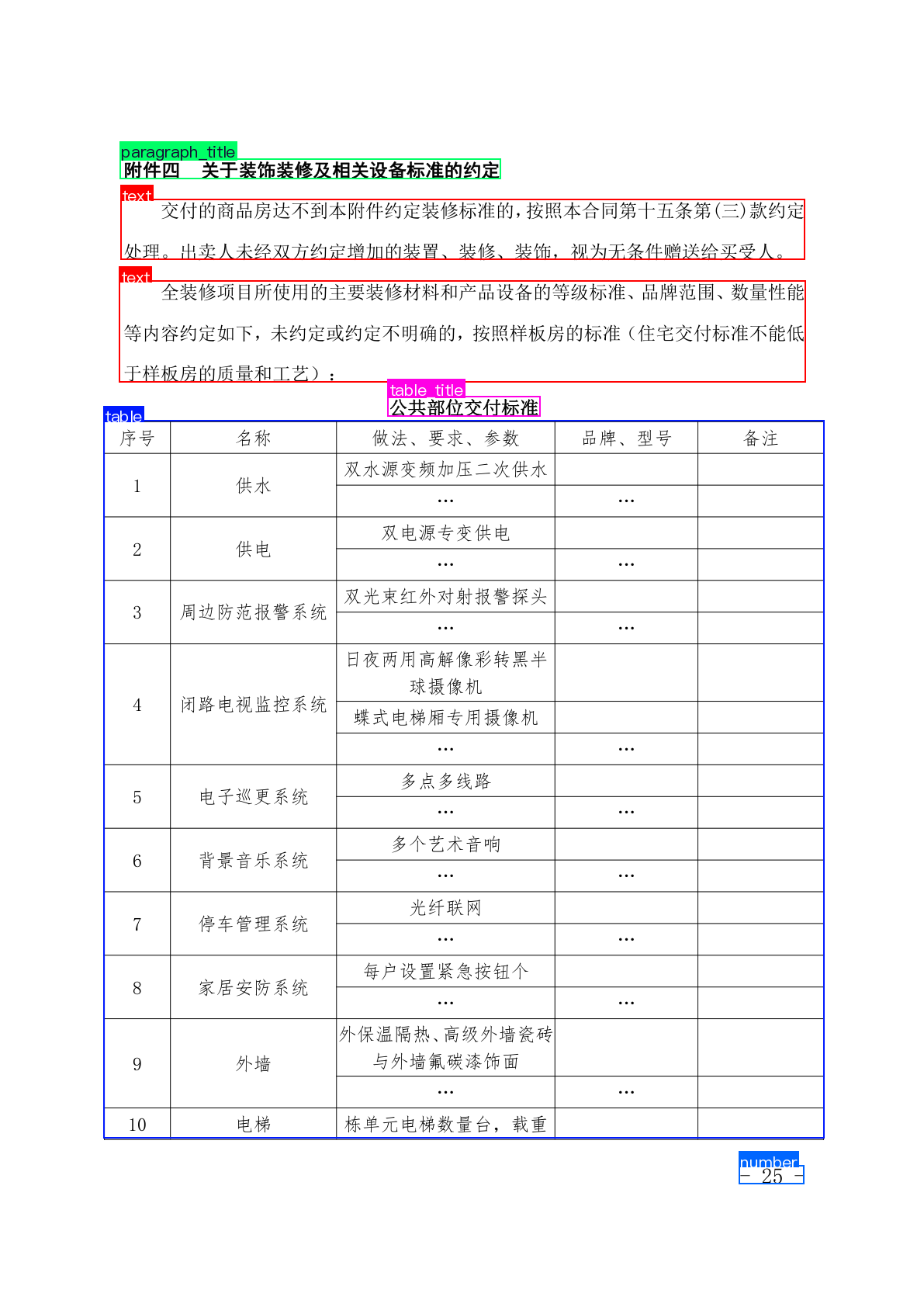}
        \end{minipage}
        \caption{Books, Notes， Test Paper and Contracts Visualization}
    \end{subfigure}
    \caption{Different visualizations by PP-DocLayout-L for various document types and layout structures.}
    \label{fig:visualizations}
\end{figure*}

We provide additional visualization results to further demonstrate the effectiveness of our model across a diverse range of document types and layouts as shown in Figure \ref{fig:visualizations}. Specifically, we visualize the performance of our PP-Layout-L model on documents such as \textit{papers}, \textit{magazines}, \textit{newspapers}, \textit{research reports}, \textit{books}, \textit{notebooks}, \textit{contracts} and \textit{test papers}. These documents exhibit varying layouts, including multi-column structures, complex headers and footers, embedded formulas, tables, and images. The visualizations clearly show that our model accurately identifies and categorizes diverse elements, such as titles, paragraphs, formulas, and figures, while maintaining robust performance across different document types. This highlights the versatility and generalization capability of our approach in handling real-world document analysis tasks.

\end{document}